\documentclass[11pt]{article}
\usepackage[T1]{fontenc}
\usepackage[utf8]{inputenc}
\usepackage{arxiv}                       
\usepackage{microtype}
\usepackage{amsmath,amssymb,bm}
\usepackage{booktabs}
\usepackage{graphicx}
\usepackage{tabularx}
\usepackage{makecell}
\usepackage{array}
\usepackage[numbers,sort&compress]{natbib}
\usepackage{url}
\usepackage{xcolor}
\usepackage[hidelinks]{hyperref}         

\graphicspath{{figures/}}
\newcommand{\X}{\mathbf{X}}
\newcommand{\Y}{\mathbf{Y}}
\newcommand{\A}{\mathbf{A}}
\newcommand{\K}{\mathbf{K}}
\newcommand{\B}{\mathbf{B}}
\newcommand{\Z}{\mathbf{Z}}
\newcommand{\T}{\mathbf{T}}
\newcommand{\R}{\mathbb{R}}
\newcommand{\transpose}{^{\mathsf T}}
\newcommand{\code}[1]{\nolinkurl{#1}}

\renewcommand{\shorttitle}{Operator-adaptive PLS and Ridge for NIR calibration}

\title{Reframing preprocessing selection as model-internal calibration in near-infrared spectroscopy: A large-scale benchmark of operator-adaptive PLS and Ridge models}

\author{
  Gr\'egory Beurier$^{1,2}$\thanks{Corresponding author: \texttt{gregory.beurier@cirad.fr}. This arXiv preprint is the extended version of a companion study prepared as a concise journal article; the two share the same results but differ in framing, structure and scope.}
  \quad Robin Reiter$^{1,2}$
  \quad Camille No{\^u}s$^{3}$ \\[0.3em]
  Lauriane Rouan$^{1,2}$
  \quad Denis Cornet$^{1,2}$ \\[0.6em]
  {\normalfont\normalsize $^{1}$CIRAD, UMR AGAP Institut, F-34398 Montpellier, France}\\
  {\normalfont\normalsize $^{2}$UMR AGAP Institut, Univ Montpellier, CIRAD, INRAE, Institut Agro, Montpellier, France}\\
  {\normalfont\normalsize $^{3}$Laboratoire Cogitamus, \url{https://www.cogitamus.fr/}}
}
\date{\today}

\begin{document}
\maketitle

\begin{abstract}
Preprocessing screening is often the most expensive part of a near-infrared
spectroscopy calibration workflow. It works because smoothing, derivatives,
detrending and related filters change the spectral directions seen by partial
least squares (PLS) or Ridge regression, but a full external search repeatedly
refits nearly the same linear model. This paper studies the case where that
search can be collapsed into one calibration step. For a strict linear
preprocessing operator $\A$ acting on row spectra as $\X\A\transpose$, the
transformed PLS cross-covariance satisfies
$(\X\A\transpose)\transpose\Y=\A\X\transpose\Y$, and Ridge regression depends on
the operator-induced kernel $\X\A\transpose\A\X\transpose$. These identities let a
finite operator bank be screened \emph{inside} the model while retaining
original-wavelength coefficients, and the same identity extends to cheaply
evaluated linear operator chains. Sample-adaptive or fitted corrections such as
SNV, MSC, EMSC and ASLS are not strict linear; we prove the boundary and keep
them as fold-local branches, not absorbed into the algebra. The study uses the
AOM benchmark cohort: 61 regression rows and 17 classification rows in the
manifest, with a strict paired regression denominator of $N=32$ rows for the
eight paper variants. On that denominator, the plain AOM-PLS (simple) records
median RMSEP ratios of 0.991 against PLS-default and 0.990 against PLS-HPO, and
the ASLS-branch AOM-PLS (best) records 0.985 and 1.002 on the same two
references. The plain AOM-Ridge (simple) records 0.974 against Ridge-default and
0.984 against Ridge-HPO, while the blended AOM-Ridge (best) records 0.918 and
0.966. The operator-adaptive classifier AOM-PLS-DA improves balanced accuracy by
a median 0.159 on $N=13$ datasets, with $12/13$ wins. The runtime gap is the
practical result: PLS-HPO takes a median total time of 710.81\,s per run, whereas
AOM-PLS takes 1.18--1.63\,s --- 436 to 602 times less PLS fitting time. Linear
operator-adaptive calibration therefore gives prediction quality comparable to
exhaustive preprocessing screening, with orders-of-magnitude less fitting time
for PLS.
\end{abstract}


\section{Introduction}
Near-infrared spectroscopy (NIRS) is a workhorse of analytical chemistry: it is
fast, non-destructive and easy to run in routine settings
\citep{burns2007handbook,pasquini2018near}. In a NIRS calibration the regressor
is rarely the bottleneck --- the \emph{preprocessing} placed in front of it is.
Raw spectra that defeat a model often become tractable after a derivative, a
smoother or a baseline correction, and a pipeline tuned on one instrument or
sample family can lose accuracy on another once the preprocessing no longer fits.
Most of the calibration effort therefore goes into screening normalisations,
smoothers, derivative orders, baselines and the number of latent variables
\citep{rinnan2009review,engel2013breaking}, and an ill-chosen preprocessing can
itself worsen a NIR model \citep{mishra2021preproc}.

That screening is almost always run \emph{outside} the model: a grid of
preprocessing recipes and model settings is enumerated, cross-validated, and the
best combination retained. The grid is large and far from independent --- in the
benchmark studied here a conventional preprocessing grid holds 600 combinations,
which under hyperparameter optimisation becomes 3000 PLS and 6000 Ridge
calibrations per dataset and seed before the final refit. The cost is not only
compute. On the small calibration sets typical of NIRS, crowning the winner of a
large inner comparison spends calibration variance and inflates apparent
accuracy --- the well-documented selection-bias regime of chemometrics and
statistics \citep{varma2006bias,cawley2010overfitting,bergstra2012random}.

\begin{figure}[t]
  \centering
  \includegraphics[width=\linewidth]{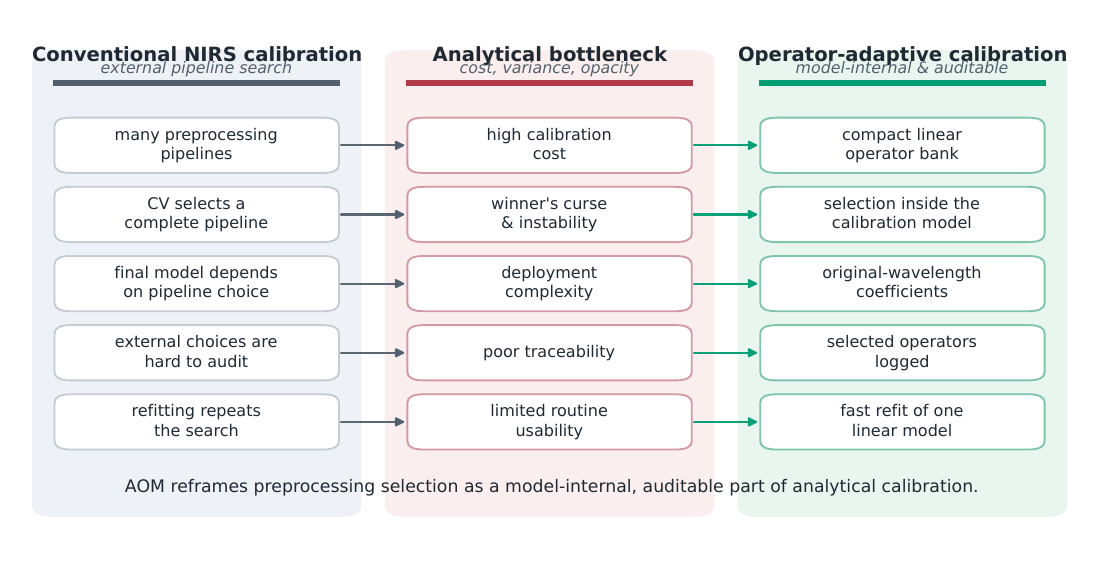}
  \caption{External preprocessing comparison repeatedly fits transformed
  pipelines (left). Operator-adaptive calibration moves the strict linear
  alternatives inside the calibration model (right); fitted or sample-adaptive
  corrections remain fold-local branches. The deployed object is a single
  linear calibration whose coefficients live on the original wavelength grid.}
  \label{fig:concept}
\end{figure}

The premise of this paper is that much of that grid never needed to leave the
model. A large share of the conventional recipes are \emph{fixed linear operators
on the wavelength axis} --- Savitzky--Golay smoothing
\citep{savitzky1964smoothing} and its derivatives, Norris--Williams gap
derivatives \citep{norris1976influence}, finite differences, polynomial
detrending. When such an operator is a fixed matrix, screening it externally
merely refits the same linear calibration through a different linear lens. We
show the linear part of the choice can instead be folded into a single PLS, a
single Ridge, or a single cheaply-screened operator-chain calibration, leaving
one linear model whose coefficients stay on the original wavelength grid
(Fig.~\ref{fig:concept}). The rest of the paper measures how far this reframing
carries on a large, heterogeneous NIR benchmark.

\paragraph{Contributions.} (i) A \emph{large-scale benchmark} of the
operator-adaptive construction on a heterogeneous cohort: an operator-adaptive
PLS-DA classification gain of 0.159 balanced accuracy on $N=13$ tasks ($12/13$
wins, Holm $p=0.007$), fixed-recipe regression gains of $2.0\%$ and $5.1\%$ median
RMSEP ($31/53$ and $34/52$ wins), a significant AOM-Ridge improvement over default
Ridge (0.974, $25/32$, Holm $p=0.007$), and AOM-PLS at parity with the full
preprocessing search (PLS-HPO; 0.990--1.002) for 436--602 times less PLS fitting
time (Section~\ref{sec:results}). (ii) The \emph{algebraic identities} that move a
strict-linear operator bank inside one PLS calibration, one Ridge calibration and
one operator-chain calibration, with the coefficients recovered on the original
wavelength axis (Sections~\ref{sec:pls}--\ref{sec:fast}), backed by auxiliary
selection rationales (Sections~\ref{sec:soft}--\ref{sec:winner}) and a numerical
check that the folding is exact rather than approximate
(Section~\ref{sec:validation}). (iii) A \emph{scope boundary} that separates the
strict-linear operators from the canonical scatter and baseline corrections (SNV,
MSC, EMSC, ASLS), which are provably not strict linear and therefore stay
fold-local (Section~\ref{sec:scope}). (iv) A public implementation,
\texttt{nirs4all-aom} (Section~\ref{sec:library}).

The scope is deliberately narrow: operator-adaptive calibration targets fixed
linear wavelength operators, while fitted or sample-adaptive scatter and baseline
corrections stay fold-local branches --- and the failure-mode analysis flags the
datasets where those corrections, not the linear bank, carry the calibration.

\section{Related work}
\label{sec:related}
Spectral preprocessing is central to chemometrics, and reviews stress that no
single recipe wins across datasets, so a practical workflow must choose among a
family of candidates, usually through an outer comparison
\citep{rinnan2009review,engel2013breaking,roger2022aquaphotomics}. A productive
line of work \emph{combines} several preprocessed views rather than selecting
one: Sequential Preprocessing through Orthogonalization
\citep[SPORT,][]{roger2020sport}, Parallel Orthogonalized PLS
\citep[PORTO,][]{mishra2021porto} and response-oriented sequential alternation
\citep[PROSAC,][]{mishra2022prosac}, and stacking plus ridge regression for
ensemble preprocessing \citep[SPRR,][]{huang2024sprr}, within the broader
ensemble framing of multiple preprocessing techniques \citep{mishra2020ensemble}.

These methods keep preprocessing \emph{outside} the model and fuse the
alternatives downstream through multiblock PLS. The approach studied here is
complementary and, for the linear part of the choice, structurally different:
strict linear preprocessings are treated as fixed operators selected
\emph{algebraically inside} one PLS or Ridge calibration, yielding a single
deployable linear model whose coefficients live on the original wavelength
grid. Table~\ref{tab:differentiation} states the contrast a chemometrics
reviewer will test.

\begin{table}[t]
  \centering
  \caption{Where operator-adaptive calibration sits relative to
  preprocessing-ensemble methods.}
  \label{tab:differentiation}
  \small
  \begin{tabularx}{\linewidth}{p{0.26\linewidth}XX}
  \toprule
  & Preprocessing ensembles (SPORT/PORTO/PROSAC) & Operator-adaptive (this work) \\
  \midrule
  Preprocessing location & outside the model & inside the calibration \\
  Combination mechanism & multiblock fusion of separately fitted blocks & algebraic selection within a fixed linear operator bank \\
  Number of fitted models & several preprocessed blocks & one calibration \\
  Deployed object & block weights + downstream model & one linear model, coefficients on original wavelengths \\
  Scatter/baseline (SNV/MSC/ASLS) & as preprocessed blocks & explicit fold-local branches (out of the linear scope) \\
  \bottomrule
  \end{tabularx}
\end{table}

This internal operator screening is complementary to multiblock preprocessing
ensembles: SPORT, PORTO and PROSAC can exploit fitted or sample-adaptive blocks
that are deliberately outside the strict-linear scope of AOM. The present
comparison is therefore against default PLS/Ridge, a fixed conventional recipe
and a full conventional preprocessing search under HPO; an empirical multiblock
comparison is a separate study.

PLS remains the reference linear method for NIRS \citep{geladi1986partial,wold2001pls},
with NIPALS and SIMPLS giving score-based and covariance-based views
\citep{dejong1993simpls}. Ridge regression provides a complementary regularized
linear model whose dual depends on the sample Gram matrix, so preprocessing acts
as a change of kernel geometry \citep{hoerl1970ridge,scholkopf2002learning,hastie2009elements}.
We combine these two views under one operator-adaptive construction.

\section{Methods}
\label{sec:methods}
Let $\X\in\R^{n\times p}$ be a centered spectral matrix (rows: samples; columns:
wavelengths) and $\Y$ the centered response. We write $\mathbf{S}=\X\transpose\Y$
for the cross-covariance. The construction below is a method principle rather
than a new theory of PLS or Ridge. PLS searches for latent directions of $\X$
that covary with $\Y$, so the cross-covariance $\mathbf{S}$ is the object that
drives the predictive directions. Ridge regression, in its dual form, depends on
the geometry of samples through the Gram matrix $\X\X\transpose$. A fixed linear
preprocessing changes exactly these two objects: it left-multiplies the PLS
cross-covariance and changes the Ridge kernel. Operator-adaptive calibration
uses this fact to move the strict-linear part of preprocessing selection inside
the calibration instead of evaluating it as an external pipeline grid.

\subsection{Scope of strict-linear and fold-local preprocessing}
\label{sec:scope}
A preprocessing is a \emph{strict linear operator} if it is a fixed matrix
$\A\in\R^{p\times p}$ acting on row spectra as $\X_{\A}=\X\A\transpose$, where
$\A$ does not depend on the response, the validation fold, the current sample,
or a reference spectrum estimated from the calibration set. Identity,
Savitzky--Golay smoothing and derivatives, finite differences, Norris--Williams
gap derivatives and polynomial detrending projections are strict linear.
Standard normal variate (SNV), multiplicative scatter correction (MSC),
extended MSC (EMSC) and asymmetric least-squares baselines (ASLS) are
\emph{not}: SNV rescales each spectrum by its own standard deviation, MSC/EMSC
fit parameters against a reference, and ASLS is an iterative fit.
The boundary is the load-bearing honesty of the method, not a caveat: the
algebra below applies exactly to the strict linear bank and provably not to the
fitted corrections, which therefore stay fold-local branches. The compact bank
used throughout is small by design (Table~\ref{tab:operators}); the rationale
for keeping it small is Section~\ref{sec:winner}.

\begin{table}[t]
  \centering
  \caption{Operator-bank scope. Strict linear operators are handled by the
  identities below; fitted or sample-adaptive corrections remain explicit
  fold-local branches.}
  \label{tab:operators}
  \small
  \begin{tabularx}{\linewidth}{p{0.23\linewidth}X p{0.24\linewidth}}
\toprule
Family & Operators in compact bank & Strict linear status \\
\midrule
Identity & identity transform & yes \\
Smoothing & Savitzky--Golay smoothing, windows 11 and 21 & yes \\
Derivatives & Savitzky--Golay first derivative, windows 11 and 21 & yes \\
Derivatives & Savitzky--Golay second derivative, window 11 & yes \\
Baseline trend & polynomial detrending, degrees 1 and 2 & yes, fixed projection on the wavelength grid \\
Local contrast & first finite difference & yes \\
\midrule
Scatter correction & SNV, MSC, EMSC & branch preprocessing, fold-local \\
Baseline correction & ASLS and related asymmetric smoothers & branch preprocessing, fold-local \\
\bottomrule
\end{tabularx}

\end{table}

\subsection{AOM-PLS: acting on the PLS cross-covariance}
\label{sec:pls}
\begin{figure}[t]
  \centering
  \includegraphics[width=\linewidth]{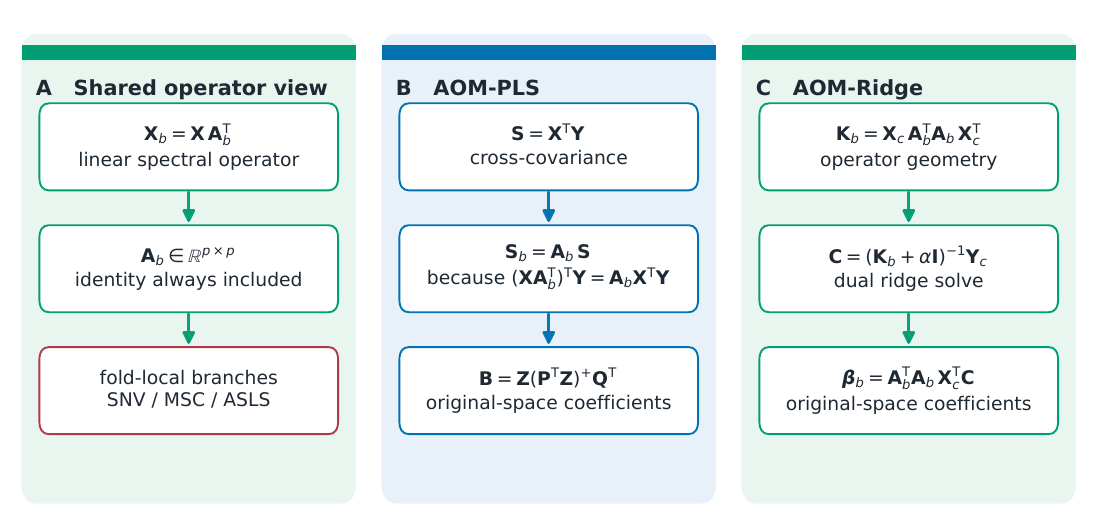}
  \caption{Operator-adaptive structure. AOM-PLS screens operators through
  cross-covariances ($\mathbf{S}_b=\A_b\mathbf{S}$); AOM-Ridge screens
  operator-induced kernels ($\K_b=\X_c\A_b\transpose\A_b\X_c\transpose$). Both
  recover coefficients on the original wavelength grid.}
  \label{fig:math}
\end{figure}

For centered $\X,\Y$ and a strict linear operator $\A_b$, applying the operator
to the spectra gives $\X_b=\X\A_b\transpose$. The cross-covariance used by PLS
then becomes
\begin{equation}
  \mathbf{S}_b=\X_b\transpose\Y=(\X\A_b\transpose)\transpose\Y
  =\A_b\,\X\transpose\Y=\A_b\mathbf{S}.
  \label{eq:cov_identity}
\end{equation}
Equation~\eqref{eq:cov_identity} is the central simplification. The PLS
cross-covariance step depends on $\X$ only through
$\mathbf{S}=\X\transpose\Y$, so a whole bank $\{\A_b\}$ is evaluated by
left-multiplication, $\mathbf{S}_b=\A_b\mathbf{S}$, at cost
$O(p\,q)$ per structured (banded or low-rank) operator --- $O(p^2 q)$ if dense ---
instead of the $O(n\,p)$ needed to materialise $\X_b=\X\A_b\transpose$.
NIPALS-adjoint and SIMPLS-covariance are implementation routes used to validate
the same method principle, not separate models
(Figure~\ref{fig:math}; Section~\ref{sec:validation}).

\subsection{AOM-Ridge: the operator-induced kernel}
\label{sec:ridge}
For centered $\X_c,\Y_c$, regularization $\alpha>0$ and a strict linear operator
$\A_b$, Ridge regression in the dual depends on the preprocessed spectra only
through the operator-induced kernel
\begin{equation}
  \K_b=(\X_c\A_b\transpose)(\X_c\A_b\transpose)\transpose
  =\X_c\A_b\transpose\A_b\X_c\transpose,
  \label{eq:ridge_kernel}
\end{equation}
with $\mathbf C=(\K_b+\alpha\mathbf I_n)^{-1}\Y_c$. As with PLS, the operator
enters the calibration through the algebraic object used by the model rather
than through a separately materialised preprocessing pipeline.

\subsection{Original-wavelength coefficients and deployment}
\label{sec:deployment_coefficients}
The main analytical consequence is that the selected model is a single linear
calibration expressed on the original wavelength grid. In AOM-PLS, a
transformed-space direction $r_a$ selected under operator $\A_b$ maps back
through the adjoint, $z_a=\A_b\transpose r_a$. With
$\Z=[z_1,\dots,z_k]$ and $\T=\X\Z$, the coefficient matrix is
\begin{equation}
  \B=\Z(\mathbf P\transpose\Z)^{+}\mathbf Q\transpose .
  \label{eq:aompls_coef}
\end{equation}
In AOM-Ridge, the dual solution for operator $\A_b$ gives original-space
coefficients
\begin{equation}
  \bm\beta_b=\A_b\transpose\A_b\X_c\transpose
  (\K_b+\alpha\mathbf I_n)^{-1}\Y_c .
  \label{eq:aomridge_coef}
\end{equation}
A weighted Ridge operator mixture uses
$\K=\sum_b s_b^2\,\X_c\A_b\transpose\A_b\X_c\transpose$ and recovers the
matching original-space coefficient through
$\sum_b s_b^2\,\A_b\transpose\A_b$, without materialising the wide block matrix.
Thus no strict-linear preprocessing sequence is replayed at prediction time:
deployment is a dot product between the incoming spectrum and the fitted
coefficient vector, plus the usual centering/intercept bookkeeping.

\subsection{FastAOM: algorithmic extension to operator chains}
\label{sec:fast}
A chain $\A_s=\A_{d}\cdots\A_{1}$ of strict linear operators composes into a
single fixed matrix, so the cross-covariance identity extends:
\begin{equation}
  (\X\A_s\transpose)\transpose\Y=\A_s\X\transpose\Y .
  \label{eq:chain_identity}
\end{equation}
This lets a large space of linear operator \emph{chains} be evaluated on
$\mathbf{S}$ at the cost of matrix--vector products, scored by an adjoint-only
Cauchy--Schwarz criterion in $[0,1]$, with denominators approximated from a
truncated singular value decomposition. The surviving chains are combined by a
\emph{sparse, non-negative (NNLS) weighting} and fit as a PLS-then-Ridge
calibration, so the result remains a linear combination of strict-linear
operator chains.  FastAOM is the demonstration that the same algebra
scales to a large linear-chain space at low fitting cost, not a separate method.

\subsection{Auxiliary selection rationale: hard operator choice}
\label{sec:soft}
The compact PLS selector uses a hard operator choice rather than a soft gate.
For a limited covariance-scoring rationale, let
$\A_\alpha=\sum_b \alpha_b\A_b$ with $\alpha$ on the simplex
$\{\alpha_b\ge0,\ \sum_b\alpha_b=1\}$. The objective
$\alpha\mapsto\lVert\A_\alpha\mathbf{S}\rVert^2$ can be written as
$\alpha\transpose\mathbf G\alpha$ with positive-semidefinite $\mathbf G$, so it
is convex and admits a maximum at a simplex vertex. Under this pure covariance
objective a soft operator mixture can therefore reduce to a single hard
operator, which explains why softmax/sparsemax gates show no advantage
\emph{under that objective}. It does not claim that soft mixtures collapse
under every criterion, nor that the cross-validated hard policy used in the
experiments is globally optimal.

\subsection{Auxiliary selection rationale: compact operator bank}
\label{sec:winner}
Enlarging the bank does not help indefinitely. More candidate operators and
component counts increase the chance of selecting a model that looks best in
validation because of noise rather than because it is truly better. The
extreme-value calculation is kept in the Supplement, because its independence
assumptions are only heuristic for correlated NIRS operators; the operational
consequence is the point here. A compact nine-operator bank screens about
$135$ operator--component candidates, whereas a hundred-operator bank screens
about $1500$, and the larger bank did not improve held-out accuracy. We
therefore use the compact bank as a prudent methodological choice: rich enough
to cover common strict-linear transformations, but small enough to limit
selection variance (Table~\ref{tab:operators}).

\subsection{Implementation and numerical equivalence}
\label{sec:validation}
For fixed folds and fixed strict-linear operators, the folded computation is not
an approximation of the explicit transformed-space search: it is the same
computation. Covariance-space AOM-PLS and the explicit transformed-space PLS
then select the same direction and return the same coefficients to machine
precision. (The fitted corrections and the truncated-SVD chain selection of
Section~\ref{sec:fast} are approximate and are excluded from this exactness
claim.) On disk, the
covariance-space and adjoint-NIPALS paths agree to a maximum absolute RMSEP
difference of $3.6\times10^{-11}$ across 159 dataset--seed pairs; a materialised
reference matches the fast paths to $10^{-6}$; and an independent C++ engine
reproduces the operator index, component count and prediction RMSEP exactly
(absolute RMSEP difference $0.000$).  For the strict-linear operators this is direct evidence that
selecting operators inside the calibration recovers exactly what the explicit
transformed-space computation would have found.

\subsection{Software implementation}
\label{sec:library}
The \texttt{nirs4all-aom} Python package is the \textbf{reference implementation}
for this paper: sklearn-compatible \texttt{fit}/\texttt{predict} estimators for
operator-adaptive PLS, Ridge and the chain selector, a one-command synthetic
smoke test, and inspectable original-wavelength coefficients. The same numerical
methods are being ported to a portable C\texttt{++} implementation with
multi-language bindings in \texttt{nirs4all-methods} (and its \texttt{pls4all}
subset), complementary to --- not a replacement for --- the Python reference; the
C\texttt{++} selection core already matches the Python reference to machine
precision (Section~\ref{sec:validation}).  Validation
status is summarized in Table~\ref{tab:software}.

\begin{table}[t]
  \centering
  \caption{Software artifacts and validation status.}
  \label{tab:software}
  \small
  \begin{tabularx}{\linewidth}{p{0.30\linewidth} X p{0.20\linewidth}}
\toprule
Component & Tests / evidence & Status \\
\midrule
\texttt{nirs4all-aom} Python package & AOM-PLS, POP-PLS, AOM-Ridge and FastAOM implementations, with unit tests and sklearn-compatible APIs. & reference implementation \\
\texttt{nirs4all-methods} / \texttt{pls4all} (C\texttt{++}) & Portable C\texttt{++} / multi-language port of the selection core; matches the Python reference to machine precision. & complementary, forthcoming \\
\texttt{nirs4all-aom} benchmark artifacts & Benchmark runners, result CSVs, aggregation scripts, figure builders and manuscript tables used here. & release with paper \\
\texttt{nirs4all} instrumentation context & NIRS instrumentation, acquisition and provenance context for local benchmark inputs. & instrumentation software \\
Supplementary validation dossier & Cohort manifest, missing-dataset audit, paired statistics and software-readiness notes distributed with \texttt{nirs4all-aom}. & public evidence \\
\bottomrule
\end{tabularx}

\end{table}

\section{Datasets and protocol}
\label{sec:data}
\subsection{Cohort}
\label{sec:cohort}
\begin{figure}[t]
  \centering
  \includegraphics[width=0.92\linewidth]{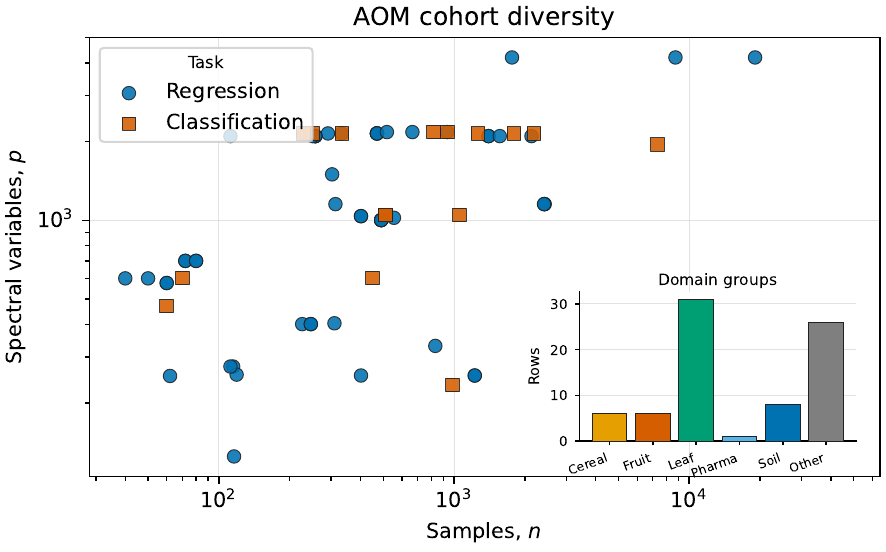}
  \caption{Cohort diversity. Sample count versus number of spectral variables on
  logarithmic axes; colours separate regression and classification rows.}
  \label{fig:dataset_diversity}
\end{figure}

The benchmark cohort is intentionally heterogeneous: 61 regression and 17
classification NIR tasks spanning leaf physiology, fruit quality, grain and seed
traits, dairy, beverages, meat quality, petroleum, soil and manure, wood
products and pharmaceutical tablets. A row denotes a concrete prediction or
classification task. Regression rows have median $n=402$ samples and median
$p=1023$ spectral variables; classification rows have median $n=511$,
$p=1951$, two classes and a 0.513 largest-class share
(Table~\ref{tab:dataset_statistics}, Figure~\ref{fig:dataset_diversity}).

\begin{table}[t]
  \centering
  \caption{Dataset shape summary for the cohort. $n$ is the total calibration
  plus external-test sample count, $p$ the number of spectral variables, $C$ the
  number of classes and $I$ the largest-class share.}
  \label{tab:dataset_statistics}
  \small
  {\setlength{\tabcolsep}{4pt}
\begin{tabular}{lrrrrrrrrrr}
\toprule
Task & $N$ & $n_\text{median}$ & $n_{\min}$ & $n_{\max}$ & $p_\text{median}$ & $p_{\min}$ & $p_{\max}$ & $\left(p/n\right)_\text{median}$ & $C_\text{median}$ & $I_{\text{median}}$ \\
\midrule
Classification & 17 & 511 & 56 & 7323 & 1951 & 235 & 2177 & 2.055 & 2 & 0.513 \\
Regression & 61 & 402 & 40 & 45417 & 1023 & 125 & 4200 & 2.382 & -- & -- \\
\bottomrule
\end{tabular}}

\end{table}

\subsection{Reference methods and comparison budgets}
\label{sec:baselines}
Comparisons use \emph{PLS and Ridge references only}. The untuned references are
PLS and Ridge with five-fold cross-validated component/regularization selection
(PLS-default, Ridge-default). The strong references are a full conventional
preprocessing comparison under hyperparameter optimisation (PLS-HPO, Ridge-HPO) over
the cartesian space
$\textit{norm}\times\textit{smooth}\times\textit{baseline}\times\textit{osc}\times n_{\text{comp}}$,
with $\textit{norm}\in\{\text{none},\text{SNV},\text{MSC},\text{EMSC}_2\}$,
Savitzky--Golay and Gaussian smoothers, $\textit{baseline}\in\{\text{none},\text{detrend},\text{ASLS}\}$
and oversampling-correction stages. The grid contains 600 preprocessing
combinations, evaluated with 5 cross-validated trials per combination for PLS
and 10 for Ridge, i.e.\ 3000 and 6000 calibration fits per dataset and seed. This
space already contains the conventional SNV $+$ Savitzky--Golay $+$ first
derivative recipe. Aggregating the configurations actually selected across
datasets shows that no single recipe dominates: the most frequently chosen
preprocessing is SNV with detrending for PLS and SNV with a Gaussian smoother for
Ridge, but each accounts for under 10\% of fits and the choice is strongly
dataset-dependent (see the Supplement). The de-facto fixed
recipe is therefore weak --- consistent with the long-standing observation that
no preprocessing wins across datasets, which is exactly what operator-adaptive
selection internalises.
A literally fixed single recipe applied uniformly across the cohort (per-sample
SNV $+$ first-derivative Savitzky--Golay, with $n_{\text{comp}}$ and $\alpha$ by
five-fold cross-validation) confirms this contrast directly: operator-adaptive
AOM-PLS and AOM-Ridge improve on the same fixed recipe by $2.0\%$ and $5.1\%$ in
median RMSEP and win the majority of datasets ($31/53$ and $34/52$), while the
fixed recipe itself yields no systematic gain over plain PLS or Ridge (median
RMSEP ratios $1.00$ and $1.02$; see the Supplement).

\subsection{Splits, selection and statistics}
\label{sec:protocol}
Where a source dataset defined an external split it was preserved; otherwise
deterministic SPXY was used for regression and stratified SPXY for
classification \citep{kennard1969computer,galvao2005method}. External test sets
were never used for preprocessing, operator, component or regularization
selection, and no outlier removal was applied. Model selection used inner
five-fold cross-validation. RMSEP is the regression metric; because response
scales differ, aggregate results use paired RMSEP ratios (a ratio below one
favours the row method). Classification uses balanced accuracy. Paired
comparisons use one-sided Wilcoxon signed-rank tests with Holm correction within
the full displayed family of comparisons, complemented by Friedman with Nemenyi
critical-difference analysis and Cliff's $\delta$; 95\% confidence intervals on
median ratios use a paired bootstrap. We report these conservative full-family
$p$-values throughout (a smaller pre-registered family would give smaller
values; see the Supplement reporting-convention note). Every comparison is
reported on its explicit paired
denominator: the strict intersection of cohort rows available for all eight
paper variants is $N_{\cap}=32$. The narrow full-HPO intersection is a coverage
limit, not a cherry-pick: unioning the three tuned-linear protocols already
available (full-HPO, model-only default-CV5, and externally-tuned PLS/Ridge)
covers 59 of the 61 regression datasets for both PLS and Ridge --- the two
exceptions being FUSARIUM targets that fail every linear method with non-finite
inputs (see the Supplement). We keep the headline paired tests
on the strict full-HPO intersection for protocol consistency and report the wider
coverage as a robustness check. The Supplement also reports a representativity
audit of the strict subset (32 of 61 regression rows; 15 of 25 source families
and 10 of 17 regression domains) and the largest pairwise denominator available
for each comparison. Thus the $N_{\cap}=32$ analysis is the protocol-consistent
headline, whereas the wider pairwise tables are sensitivity checks rather than
new headline claims. The AOM-Ridge calibrations are deterministic, so
across-seed variance is essentially zero by construction rather than by
small-sample luck (see the Supplement); the relevant robustness
axis is the data partition. This study contains one held-out-site transfer check
(Section~\ref{sec:transfer}); it does not replace a repeated random-partition
benchmark.

\begin{table}[t]
  \centering
  \caption{Cohort denominators and data-quality rules used in the main
  manuscript.}
  \label{tab:benchmark_diversity}
  \small
  \begin{tabularx}{\linewidth}{p{0.34\linewidth}X}
\toprule
Property & Summary used in this work \\
\midrule
Regression manifest & 61 included regression rows across the benchmark inventory. \\
Classification manifest & 17 classification manifest rows; 16 included rows after one missing-file exclusion. \\
Main regression denominator & $N_{\cap}=32$ datasets for the eight paper variants. \\
Paper variants & PLS-default, PLS-HPO, AOM-PLS (simple), AOM-PLS (best), Ridge-default, Ridge-HPO, AOM-Ridge (simple) and AOM-Ridge (best). \\
Analytical domains & Leaf physiology, fruit quality, grain and seed traits, dairy, beverages, meat quality, petroleum, soil, manure, wood products, tablets and public calibration datasets. \\
Sample and wavelength range & Training sets span 28--39{,}225 samples and 125--4{,}200 spectral variables in the local regression manifest. \\
Validation rule & External test sets are never used for preprocessing, operator, component or regularization selection. \\
\bottomrule
\end{tabularx}

\end{table}

\section{Results}
\label{sec:results}

\subsection{PLS: comparable accuracy at a fraction of the budget}
\label{sec:res_pls}
On the strict regression denominator, the plain operator-adaptive PLS (AOM-PLS,
simple) reaches a median RMSEP ratio of 0.991 against PLS-default ($22/32$ wins)
and is at near parity with PLS-HPO, 0.990 ($19/32$). Adding a fold-local ASLS
branch before the compact bank (AOM-PLS, best) gives 0.985 against PLS-default
($20/32$) and 1.002 against PLS-HPO ($15/32$): \emph{comparable to}, not better
than, the exhaustive preprocessing search (Table~\ref{tab:main_results}). The
simple variant is the compact nine-operator bank selected by internal five-fold
cross-validation, with no fitted branch. These PLS effects are small and not
confirmatory under the conservative full-family Holm tests; their practical value
is that near-parity is obtained at much lower fitting cost. The
reading is deliberately plain: full preprocessing-HPO can recover the benefit of
careful preprocessing, but it does so by refitting the linear calibration under
a far larger comparison budget, whereas AOM-PLS makes the strict-linear part of the
same decision inside one coefficient-bearing model.

\begin{table}[t]
  \centering
  \caption{Main paired regression results. Ratios below one favour the row
  method over the named reference; the last column gives median RMSEP ratio,
  wins, and the one-sided Holm-adjusted $p$-value (two-sided sensitivity in the
  Supplement).}
  \label{tab:main_results}
  \begingroup
  \scriptsize
  \setlength{\tabcolsep}{3pt}
  \renewcommand{\arraystretch}{0.88}
  \begin{tabularx}{\linewidth}{p{0.31\linewidth}Xrr}
\toprule
Comparison & Evidence source & $N$ & Median RMSEP ratio / wins; $p_{\mathrm{Holm}}$ \\
\midrule
AOM-PLS (simple) vs PLS-default & AOM-PLS seeds012 / default-CV all & 32 & 0.991; 22/32; 0.896 \\
AOM-PLS (best) vs PLS-default & AOM-PLS seeds012 / default-CV all & 32 & 0.985; 20/32; 1.000 \\
AOM-PLS (simple) vs PLS-HPO & AOM-PLS seeds012 / cartesian HPO seeds012 & 32 & 0.990; 19/32; 1.000 \\
AOM-PLS (best) vs PLS-HPO & AOM-PLS seeds012 / cartesian HPO seeds012 & 32 & 1.002; 15/32; 1.000 \\
PLS-HPO vs PLS-default & cartesian HPO seeds012 / default-CV all & 32 & 0.992; 19/32; 1.000 \\
Ridge-HPO vs Ridge-default & cartesian HPO seeds012 / default-CV all & 32 & 0.962; 19/32; 1.000 \\
AOM-Ridge (simple) vs Ridge-default & AOM-Ridge headline / default-CV all & 32 & 0.974; 25/32; 0.007 \\
AOM-Ridge (best) vs Ridge-default & AOM-Ridge headline / default-CV all & 32 & 0.918; 27/32; 2.6e-04 \\
AOM-Ridge (simple) vs Ridge-HPO & AOM-Ridge headline / cartesian HPO seeds012 & 32 & 0.984; 19/32; 1.000 \\
AOM-Ridge (best) vs Ridge-HPO & AOM-Ridge headline / cartesian HPO seeds012 & 32 & 0.966; 25/32; 0.033 \\
\bottomrule
\end{tabularx}

  \endgroup
\end{table}

\subsection{Ridge: operator kernels improve accuracy}
\label{sec:res_ridge}
The core Ridge evidence is the \emph{plain} operator-adaptive Ridge (AOM-Ridge,
simple): it improves on default Ridge with a median RMSEP ratio of 0.974
($25/32$ wins, Holm-adjusted $p=0.007$) and stays close to Ridge-HPO (0.984).
This single global selection over the compact strict-linear bank uses no operator
mixing or local-neighbourhood weighting. It is the clean evidence for the
operator-kernel view: in
Ridge, preprocessing changes the regularized sample geometry, and selecting the
operator inside the kernel improves the calibration while keeping a single
deployable model.

A heavier selector variant (AOM-Ridge, best) blends out-of-fold predictions of a
candidate panel and reaches a median ratio of 0.918 against Ridge-default
($27/32$, $p=2.6\times10^{-4}$) and 0.966 against Ridge-HPO ($25/32$, one-sided
$p=0.033$). This directional edge over Ridge-HPO does not reach $0.05$ under the
two-sided sensitivity analysis ($p=0.063$; Supplement), whereas the improvement
over default Ridge remains significant two-sided ($p=5.2\times10^{-4}$).
We report it as a secondary variant rather than the headline: the gain comes
from an additional selection step rather than the core algebra, and this blended
result is currently single-seed.  The practical message is the simple baseline, which already
improves on default Ridge with a clean significance.
Per-dataset absolute figures of merit (RMSEP in original units, $R^2$ and RPD)
for the $N_{\cap}=32$ intersection are reported in the Supplement; they span
fit-for-purpose calibrations and the negative-$R^2$ failure cases identified in
Section~\ref{sec:failure}.

\subsection{FastAOM: a large linear-chain space, evaluated cheaply}
\label{sec:res_fast}
The chain selector (FastAOM) is competitive with PLS-default at low fitting cost,
with a median relative RMSEP near 1.01--1.05 depending on the chain combination.
Its value is the demonstration that a large space of strict-linear operator
chains can be evaluated and combined without an external comparison; the heavier
chain policies cost two orders of magnitude more time for no accuracy gain, which
argues for the cheap closed-form path rather than against it
(supplementary material).

\subsection{Classification}
\label{sec:res_cls}
Operator-adaptive PLS-DA improves balanced accuracy by a median 0.159 on the
$N=13$ paired classification cohort ($12/13$ wins, Holm-adjusted $p=0.007$;
Table~\ref{tab:classification_main}). This secondary result confirms that
covariance-space operator selection transfers from regression to classification
when the response encoding and scoring metric change. The probability calibration
has a NIR precedent in Talanta \citep{perezmarin2021}; here it is more nuanced
(Supplement): operator-adaptive PLS-DA also lowers the log-loss (median 1.16
versus 1.39 for PLS-DA) but is \emph{less} well-calibrated
(expected calibration error 0.32 versus 0.11), i.e.\ more accurate and more
confident but over-confident. We therefore present PLS-DA as the headline
classifier for the methodological point that the covariance-space selection
transfers, and note that the operator-adaptive \emph{Ridge} classifiers reach
comparable balanced accuracy with calibration on par with PLS-DA
(ECE $\approx0.09$--$0.11$, log-loss $\approx0.7$--$0.8$); the full calibration
table is in the supplement.

\begin{table}[t]
  \centering
  \caption{Main classification result. Positive differences favour the
  operator-adaptive classifier.}
  \label{tab:classification_main}
  \small
  \begin{tabularx}{\linewidth}{Xrrrr}
\toprule
Comparison & $N$ & Median $\Delta$ balanced acc. & 95\% CI & Wins; $p_{\mathrm{Holm}}$ \\
\midrule
AOM-PLS-DA-global-simpls-covariance vs PLS-DA & 13 & 0.159 & 0.129--0.422 & 12/13; 0.007 \\
\bottomrule
\end{tabularx}

\end{table}

\subsection{Time budget}
\label{sec:res_time}
\begin{figure}[t]
  \centering
  \includegraphics[width=\linewidth]{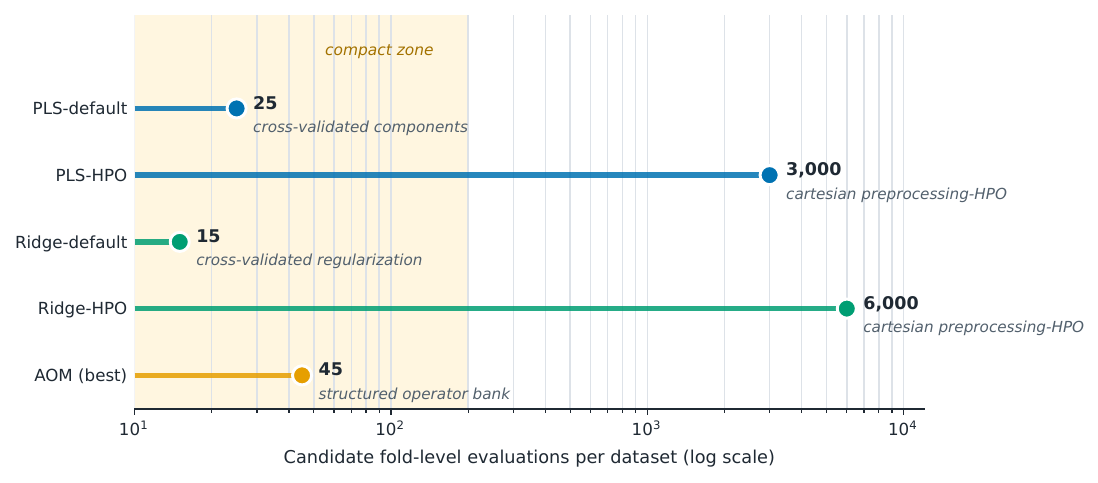}
  \caption{Comparison-budget scale (search units --- fits or evaluations per
  dataset and seed, log axis) for the eight paper variants.}
  \label{fig:budget}
\end{figure}

\begin{figure}[t]
  \centering
  \includegraphics[width=\linewidth]{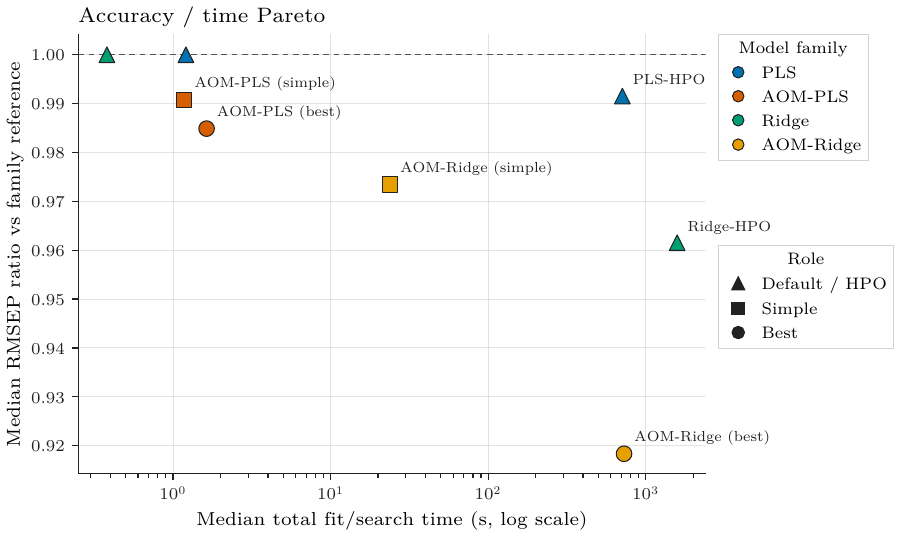}
  \caption{Accuracy/time view for the eight paper variants; the horizontal axis
  is median total fit/search time.}
  \label{fig:pareto}
\end{figure}

\begin{figure}[t]
  \centering
  \includegraphics[width=\linewidth]{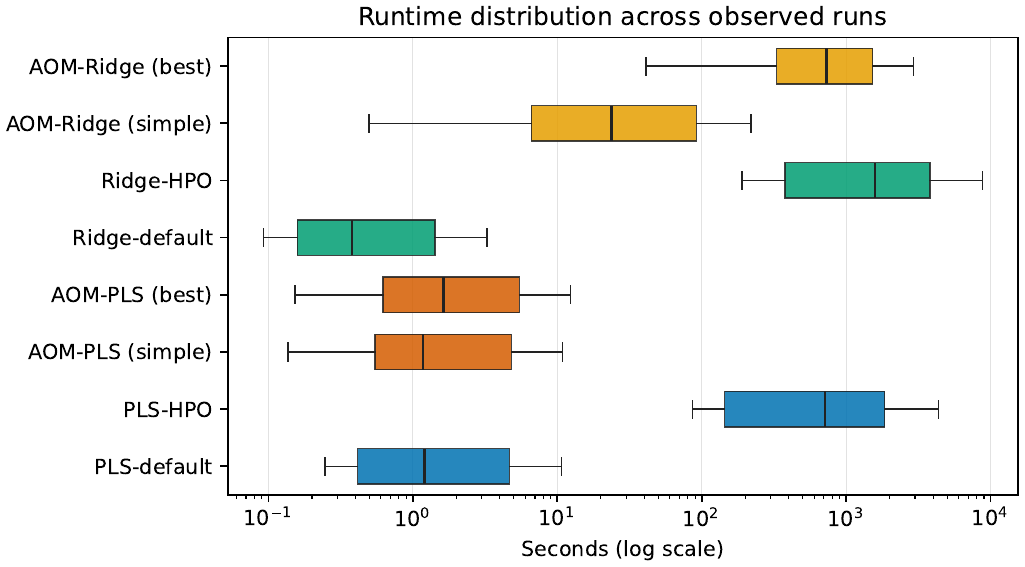}
  \caption{Runtime distributions for the eight paper variants.}
  \label{fig:runtime}
\end{figure}

The practical advantage is fitting time, and it is specific to PLS. PLS-HPO has a
median total time of 710.81\,s per run; AOM-PLS takes 1.18\,s (simple) and
1.63\,s (best), i.e.\ \textbf{436 to 602 times faster}, and reduces the
comparison from 3000 calibration fits to 45 compact-bank cross-validation fits
per dataset and seed (Table~\ref{tab:time_budget};
Figures~\ref{fig:budget}, \ref{fig:pareto} and~\ref{fig:runtime}).
The observed runtime is the primary cost evidence; the fit/evaluation counts are
scale indicators because a PLS refit, a Ridge kernel solve and a chain-screening
cell are not identical computational units.
For Ridge the picture is more modest and we state it plainly: the plain
AOM-Ridge runs in a median 23.78\,s, while the heavy blended variant takes
728.81\,s---only about $2.2\times$ faster than Ridge-HPO (1584\,s). The general
conclusion is not that every operator-adaptive selector is instant; it is that
strict-linear preprocessing can be represented as a model geometry rather than an
external preprocessing comparison, and for PLS this collapses the comparison by
two orders of magnitude.

\begin{table}[t]
  \centering
  \caption{Comparison and runtime budget for the main methods. Observed totals are
  sums over benchmark rows.}
  \label{tab:time_budget}
  \small
  \begin{tabularx}{\linewidth}{p{0.235\linewidth}rXrr}
\toprule
Method & Datasets & Search budget & Median fit (s) & Median total (s) \\
\midrule
PLS-default & 32 & 25 component trials; 0.1 h & 0.02 & 1.21 \\
PLS-HPO & 32 & 600 recipes $\times$ 5 trials; 40.8 h & 0.03 & 710.81 \\
AOM-PLS (simple) & 32 & 9 operators $\times$ CV-5; 0.1 h & 1.18 & 1.18 \\
AOM-PLS (best) & 32 & ASLS branch $+$ 9 operators $\times$ CV-5; 0.1 h & 1.43 & 1.63 \\
Ridge-default & 32 & 15 $\alpha$ trials; 0.0 h & 0.00 & 0.38 \\
Ridge-HPO & 32 & 600 recipes $\times$ 10 trials; 96.5 h & 0.05 & 1584.00 \\
AOM-Ridge (simple) & 32 & 9 operators $\times$ 50 $\alpha$ cells; 0.5 h & 23.77 & 23.78 \\
AOM-Ridge (best) & 32 & Blender: 8 candidates $\times$ 3 outer folds $+$ 8 refits; 9.3 h & 727.51 & 728.81 \\
\bottomrule
\end{tabularx}

\end{table}

\subsection{Which operators the model selects}
\label{sec:res_operators}
The selected operators are chemically meaningful and few. Across the cohort the
three most frequent strict operators (Savitzky--Golay smoothing at window 21,
and first derivatives at windows 21 and 11) account for 54.7\% of compact-bank
component selections, while identity is selected on 8.7\% of components. The bank
is diverse enough to adapt but small enough to keep selection stable
(Section~\ref{sec:winner}); and because the operator is logged and the
coefficients live on the original wavelengths, the deployed calibration is
auditable. The transfer and latency results below
(Section~\ref{sec:transfer}) make the deployment side concrete rather than
asserted.

\subsection{Transfer to held-out sites and deployment cost}
\label{sec:transfer}
On the Rd25 leaf dataset, three leave-site-out partitions (trained on two
acquisition sites, tested on a held-out third: CB, GT, XSBN) show that
operator-adaptive calibration transfers as well as the plain linear baselines:
the median RMSEP rises by a factor of 1.28 (AOM-PLS) and 1.30 (AOM-Ridge) from
the random \texttt{spxy70} split to the held-out-site split, on par with plain
PLS (1.28) and slightly below plain Ridge (1.32; Table~\ref{tab:transfer}).
Folding the preprocessing into the model therefore showed no additional transfer
penalty relative to plain PLS/Ridge in this Rd25 leave-site-out check. A
calibration that is meant to be deployed must also run cheaply at prediction
time; that cost can be read off runs already in the benchmark.
This is a deployment sanity check on one source family, not a general
instrument-transfer validation; calibration transfer and standardisation remain
active Talanta topics \citep{mishra2021dipls,xiong2024sfat}, and broader blocked
or repeated-partition studies would be needed to make a stronger robustness
claim.

The deployment cost is that of a single linear model. Median prediction time is
$0.2$--$0.6$\,ms per dataset for the operator-adaptive models, the same as plain
PLS/Ridge (Table~\ref{tab:latency}), because the fitted object is one set of
coefficients on the original wavelength grid --- there is no preprocessing
sequence to evaluate at predict time. The deployed artifact is correspondingly
small: a coefficient vector of $8p$ bytes (one double per wavelength) ---
$\approx1$--$34$\,kB across the cohort ($p=125$--$4200$; median $\approx8$\,kB) ---
plus a short text log of the selected operator, with no fitted preprocessing
transform to store or replay.

\begin{table}[t]
  \centering
  \caption{Transfer to held-out acquisition sites. Median RMSEP on three
  leave-site-out Rd25 partitions (train on two sites, test on the third) versus
  the random \texttt{spxy70} split of the same data; the gap ratio is
  blocked-site / random RMSEP.}
  \label{tab:transfer}
  \small
  \begin{tabularx}{\linewidth}{Xrrrrrr}
\toprule
Model & CB & GT & XSBN & Blocked & Random & Gap \\
 & (site) & (site) & (site) & median & (\texttt{spxy70}) & ratio \\
\midrule
AOM-PLS & 0.2288 & 0.1931 & 0.2808 & 0.2288 & 0.1788 & 1.280 \\
AOM-Ridge & 0.2255 & 0.1953 & 0.2595 & 0.2255 & 0.1741 & 1.295 \\
PLS & 0.2288 & 0.2042 & 0.3158 & 0.2288 & 0.1786 & 1.281 \\
Ridge & 0.2291 & 0.1922 & 0.2591 & 0.2291 & 0.1741 & 1.316 \\
\bottomrule
\end{tabularx}

\end{table}

\begin{table}[t]
  \centering
  \caption{Inference and fit cost on the Rd25 datasets. Operator-adaptive models
  predict in the same time as plain PLS/Ridge (one linear model, no
  preprocessing to replay).}
  \label{tab:latency}
  \small
  \begin{tabularx}{\linewidth}{Xrr}
\toprule
Model & Median predict (s) & Median fit (s) \\
\midrule
AOM-PLS & 0.0006 & 4.4785 \\
AOM-Ridge & 0.0002 & 89.8523 \\
PLS & 0.0007 & 0.1651 \\
Ridge & 0.0002 & 0.5465 \\
\bottomrule
\end{tabularx}

\end{table}

\subsection{Failure modes}
\label{sec:failure}
\begin{figure}[t]
  \centering
  \includegraphics[width=\linewidth]{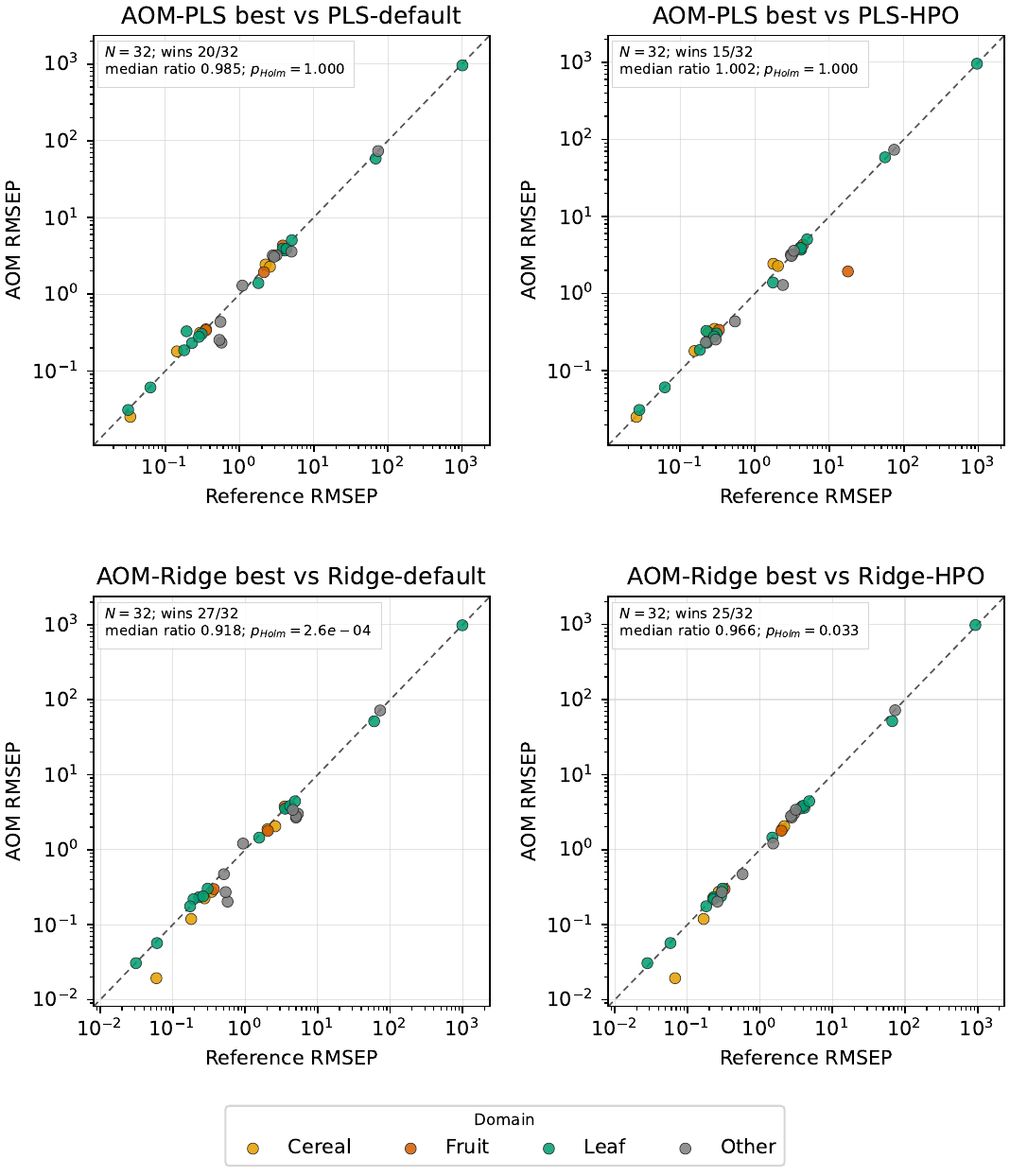}
  \caption{Prediction-quality scatter (log RMSEP axes) for the selected
  operator-adaptive variants against default and HPO references; points below
  the diagonal favour the row method.}
  \label{fig:paired_scatter}
\end{figure}

\begin{figure}[t]
  \centering
  \includegraphics[width=\linewidth]{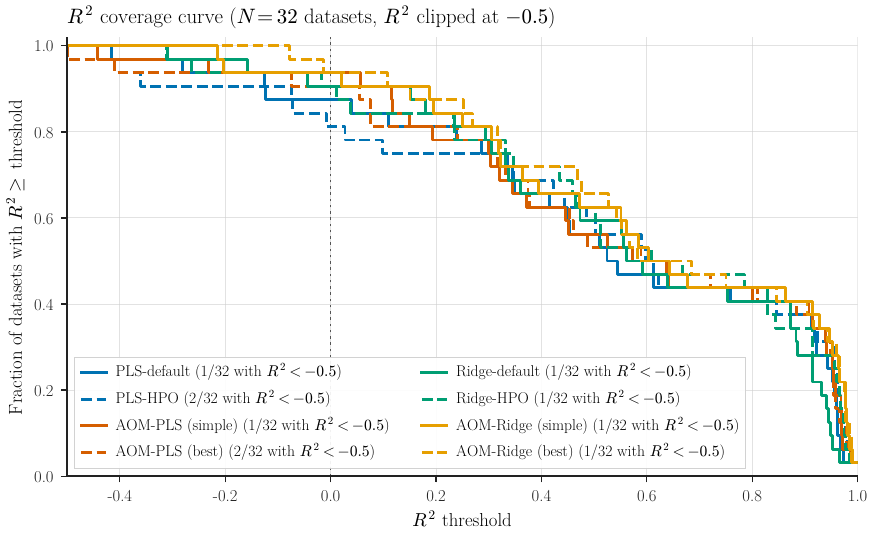}
  \caption{$R^2$ coverage curves for the eight paper variants on the main
  regression denominator (clipped at $-0.5$).}
  \label{fig:r2_cdf}
\end{figure}

The medians hide datasets where operator-adaptive PLS trails the PLS-HPO winner:
berry Brix, rice amylose, a colza nitrogen row and a strawberry-puree Brix row,
all cases where a fitted scatter or baseline correction (SNV, MSC, ASLS) carries
most of the calibration value and the strict linear bank captures only part of
it. Per-component operator selection (POP-PLS) is reported as a negative
ablation: it underperforms the global choice (median ratio 1.37--1.39), and a
local-neighbourhood Ridge variant is included as an intentional ``does not always
win'' example. These results are kept in the main text and the supplement because
honest failure modes are part of the evidence, and because they mark the
practical boundary between strict-linear preprocessing and fitted/sample-adaptive
corrections (Figure~\ref{fig:paired_scatter}, Figure~\ref{fig:r2_cdf}).

\section{Discussion}
\label{sec:discussion}
The measured advantage is compound rather than a single HPO accuracy win:
operator-adaptive PLS-DA improves balanced accuracy by 0.159 on $N=13$
classification tasks ($12/13$ wins, Holm $p=0.007$); AOM-PLS and AOM-Ridge beat
the fixed SNV $+$ first-derivative Savitzky--Golay recipe by $2.0\%$ and
$5.1\%$ median RMSEP ($31/53$ and $34/52$ wins); plain AOM-Ridge improves on
default Ridge (0.974, $25/32$, Holm $p=0.007$); and AOM-PLS remains at parity
with PLS-HPO (0.990--1.002) while using 436--602 times less PLS fitting time.

Three threads tie the result to analytical practice. \emph{Computation:} when
preprocessing alternatives are strict linear operators, comparing them outside
the model is unnecessary work---AOM-PLS folds the choice into the cross-covariance
and AOM-Ridge into the kernel, and the calibration remains linear and auditable.
\emph{Statistics:} a large cartesian grid re-estimates pipelines on small folds
and promotes the winner of a noisy comparison; the operator-adaptive step instead
selects from a covariance or kernel representation of one calibration problem and
avoids turning a derivative choice into a multiplicative refitting of the same
model (Section~\ref{sec:winner}). \emph{Deployment:} a grid winner is an external
preprocessing sequence plus a downstream model, every stage of which must be
reproduced on a new instrument or software stack, whereas AOM recovers
coefficients on the original wavelength axis
(Eqs.~\eqref{eq:aompls_coef},~\eqref{eq:aomridge_coef}),
so the deployed object is a single inspectable linear calibration. This is an
implementation and auditability advantage; the single Rd25 transfer check above
does not by itself prove broad instrument-transfer robustness. The scope is
equally clear: AOM is the right tool for the well-behaved
reflectance/absorbance regime and does not absorb sample-adaptive corrections,
which must remain fold-local and audited. Relative to SPORT/PORTO/PROSAC, the
difference is that strict linear alternatives are selected inside one calibration
rather than fused across separately fitted preprocessed blocks
(Section~\ref{sec:related}); no empirical multiblock-ensemble benchmark is
claimed here.

Finally, the cohort is clustered: its 32 paired-comparison datasets come from
only 15 source families, so dataset-level tests could overstate independence. A
source-family-clustered re-analysis (see the Supplement)
leaves the headline conclusions intact --- AOM-Ridge, the AOM-Ridge Blender and
AOM-PLS still beat their \emph{default} linear baselines in 13--14 of 15 families
($p\le0.004$) with effect sizes at least as large as at the dataset level --- but
it tempers the comparisons against the \emph{tuned} HPO baselines to $p\approx0.06$,
so those are reported as suggestive rather than confirmatory.

\section{Conclusions}
\label{sec:conclusion}
The source-traceable wins are classification, the fixed conventional recipe, and
default Ridge: operator-adaptive PLS-DA improves balanced accuracy by 0.159 on
$N=13$ tasks ($12/13$ wins, Holm $p=0.007$); AOM-PLS and AOM-Ridge beat the
fixed SNV $+$ first-derivative Savitzky--Golay recipe by $2.0\%$ and $5.1\%$
median RMSEP ($31/53$ and $34/52$ wins); and plain AOM-Ridge improves on default
Ridge (0.974, $25/32$, Holm $p=0.007$). For PLS-HPO, the result is parity and
cost rather than an accuracy win: AOM-PLS remains comparable to the conventional
preprocessing search (0.990--1.002) while using 436--602 times less PLS fitting
time. These results follow from moving strict linear preprocessing into the
calibration itself: PLS, Ridge and operator-chain identities move the operator
choice inside one model whose coefficients stay on the original wavelength grid.
The message for analytical practice is deliberately bounded: keep fitted
corrections fold-local, fold strict linear operators into the model, treat
comparisons against tuned HPO and multiblock ensembles cautiously, and report
every comparison on its paired denominator.

\section*{Data and code availability}
The software entry point for the methods, benchmark scripts, result tables and
manuscript artifacts is \texttt{nirs4all-aom} version 0.1.1
(manuscript audit commit \texttt{1dc25b1}) \citep{beurier2026aomnirs}. The
\texttt{nirs4all} project \citep{beurier2026nirs4all} is cited only for the NIRS
instrumentation, acquisition and provenance context used to assemble local
benchmark inputs. The benchmark result tables and per-run prediction CSVs are
distributed with \texttt{nirs4all-aom}; the underlying NIR spectra are third-party
datasets used under their original terms and are not redistributed here, each
identified with its source in the cohort manifest (Supplement).

\section*{Declaration of Generative AI and AI-assisted technologies in the writing process}
During the preparation of this work, the authors used Anthropic Claude and
OpenAI Codex to assist with code review, implementation, repository
organization, benchmark aggregation scripts, LaTeX editing and drafting or
revision of explanatory text.  After using these tools, the authors reviewed,
edited and verified the code, numerical results, references and manuscript
content as needed, and take full responsibility for the content of the
publication.

\section*{Supplementary material}
The supplementary material contains the full derivations, the cohort manifest,
operator-bank diagnostics, the missing-dataset reason-code audit, the
non-selected and negative AOM variants, per-dataset tables, classification
details with log-loss and calibration, representativity and pairwise-denominator
sensitivity checks, seed and split sensitivity, the numerical validation matrix
and the two-sided test sensitivity.

\bibliographystyle{unsrtnat}
\bibliography{references}

\end{document}


\maketitle
\vspace{-2.5em}
\tableofcontents
\vspace{1em}

\noindent This supplement gives the full derivations (Section~\ref{sec:s_proofs}),
the numerical-validation matrix (Section~\ref{sec:s_validation}), the cohort
manifest (Section~\ref{sec:s_cohort}), operator-bank diagnostics
(Section~\ref{sec:s_operators}), the missing-dataset reason-code audit
(Section~\ref{sec:s_missing}) and paired-denominator sensitivity
(Section~\ref{sec:s_pairwise}), the full variant families including the negative
ablations (Section~\ref{sec:s_variants}), the per-dataset results
(Section~\ref{sec:s_perdataset}), classification details
(Section~\ref{sec:s_classification}), seed and split sensitivity
(Section~\ref{sec:s_seed}), the two-sided test sensitivity
(Section~\ref{sec:s_twosided}), and the failure-mode analysis
(Section~\ref{sec:s_failure}). Numbers are pinned from the run CSVs distributed
with \texttt{nirs4all-aom}; their per-value provenance is mapped in
their on-disk source files.

\section{Derivations}
\label{sec:s_proofs}
Let $\X\in\R^{n\times p}$ be a centered spectral matrix (rows: samples; columns:
wavelengths) and $\Y\in\R^{n\times q}$ the centered response. A strict linear
operator is a fixed $\A\in\R^{p\times p}$ acting on row spectra as
$\X_{\A}=\X\A\transpose$, independent of $\Y$, of the validation fold, of the
sample, and of any reference spectrum. Write $\mathbf{S}=\X\transpose\Y$.

\subsection{Cross-covariance identity}
\begin{propositionS}[Cross-covariance identity]\label{sp:cov}
For centered $\X,\Y$ and a strict linear operator $\A_b$,
$(\X\A_b\transpose)\transpose\Y=\A_b\,\X\transpose\Y$.
\end{propositionS}
\begin{proof}
$(\X\A_b\transpose)\transpose\Y=\A_b\X\transpose\Y$ by the transpose of a
product, using that $\A_b$ is a fixed matrix independent of $\Y$. Hence
$\mathbf{S}_b\equiv(\X\A_b\transpose)\transpose\Y=\A_b\mathbf{S}$.
\end{proof}
A bank $\{\A_b\}_{b=1}^{B}$ is therefore screened on the single matrix
$\mathbf{S}$ by left-multiplication, $\mathbf{S}_b=\A_b\mathbf{S}$, at cost
$O(p\,q)$ per operator for the structured (banded or low-rank) operators of the
bank --- $O(p^2 q)$ for a dense operator --- rather than the
$O(n\,p)$ needed to materialise $\X_b=\X\A_b\transpose$ for each $b$. For
single-response PLS ($q=1$) the screened quantity is the $p$-vector
$\mathbf{s}_b=\A_b\mathbf{s}$.

\subsection{SIMPLS-covariance and NIPALS-adjoint}
SIMPLS \citep{dejong1993simpls} extracts components directly from
$\mathbf{S}$: the leading left singular vector $r=u_1(\mathbf{S})$ gives a
weight, $t=\X r$ is normalised, loadings $p=\X\transpose t$ and
$q_{\mathrm{load}}=\Y\transpose t$ follow, and $\mathbf{S}$ is deflated by
Gram--Schmidt of the loadings. Combined with Proposition~\ref{sp:cov}, each
operator candidate is scored on $\mathbf{S}_b=\A_b\mathbf{S}$ and the dominant
direction $r_b=u_1(\mathbf{S}_b)$ is mapped back to the original wavelength axis
by the adjoint, $z_b=\A_b\transpose r_b$. NIPALS may be rewritten the same way:
every step that would form $\A_b x$ is replaced by the adjoint $\A_b\transpose r$
on a residual $r$, so $\A_b$ is never materialised. Both engines coincide with
standard PLS run on the explicitly transformed matrix $\X_b$ when the operator
is fixed (verified numerically in Section~\ref{sec:s_validation}).

\subsection{Coefficient recovery on the original wavelength grid}
\begin{propositionS}[Original-grid coefficients]\label{sp:coef}
For an operator sequence $(\A_{b_1},\dots,\A_{b_K})$ with transformed-space
directions $r_a$, set $z_a=\A_{b_a}\transpose r_a$, $\Z=[z_1,\dots,z_K]$,
$\T=\X\Z$, $\PP=\X\transpose\T\,\mathrm{diag}(\lVert t_a\rVert^{-2})$ and
$\Q=\Y\transpose\T\,\mathrm{diag}(\lVert t_a\rVert^{-2})$. Then the regression
coefficient $\B=\Z(\PP\transpose\Z)^{+}\Q\transpose$ (Moore--Penrose
pseudoinverse) acts on the original (untransformed) spectra.
\end{propositionS}
\begin{proof}
Each component lives in the original space because $z_a=\A_{b_a}\transpose r_a$
and $t_a=\X z_a$ are expressed without forming $\X_b$. The expression
$\B=\Z(\PP\transpose\Z)^{+}\Q\transpose$ is the standard PLS coefficient form
$\W(\PP\transpose\W)^{+}\Q\transpose$ with the original-space weights $\Z$ in
place of $\W$. The prediction $\widehat{\Y}=\X\B+\bar{\Y}$ uses raw $\X$, so no
preprocessing stage is replayed at prediction time.
\end{proof}

\subsection{AOM-PLS in one pass: algorithm and dimensions}
The covariance-space engine is summarised below; every candidate operator is
scored by a left-multiplication of the single $p\times q$ matrix $\mathbf{S}$, so
the whole bank is screened at the cost of one PLS fit. Table~\ref{tab:s_dims}
lists the dimensions of every quantity.

\begin{table}[h]
  \centering
  \caption{Dimensions ($n$ samples, $p$ wavelengths, $q$ responses, $B$ operators,
  $K$ components).}
  \label{tab:s_dims}
  \small
  \begin{tabularx}{\linewidth}{llX}
  \toprule
  Symbol & Shape & Meaning \\
  \midrule
  $\X,\ \Y$ & $n\times p$, $n\times q$ & centered spectra, responses \\
  $\mathbf{S}=\X\transpose\Y$ & $p\times q$ & cross-covariance (the only data summary screened) \\
  $\A_b$ & $p\times p$ & strict linear operator $b=1,\dots,B$ (never materialised) \\
  $\mathbf{S}_b=\A_b\mathbf{S}$ & $p\times q$ & operator-screened cross-covariance, $O(pq)$ each \\
  $r_a$ & $p\times 1$ & transformed-space direction (leading singular vector of $\mathbf{S}_b$) \\
  $z_a=\A_{b}\transpose r_a$ & $p\times 1$ & original-grid effective weight; $\Z=[z_1\dots z_K]$ \\
  $\T=\X\Z,\ \PP,\ \Q$ & $n\times K$, $p\times K$, $q\times K$ & scores, spectral / response loadings \\
  $\B=\Z(\PP\transpose\Z)^{+}\Q\transpose$ & $p\times q$ & deployed coefficients on the original grid \\
  \bottomrule
  \end{tabularx}
\end{table}

\noindent\fbox{\begin{minipage}{0.97\linewidth}
\small\textbf{Algorithm 1.} AOM-PLS (covariance-space, global selection).\\
\textbf{Input:} centered $\X,\Y$; operator bank $\{\A_b\}_{b=1}^{B}$ (identity
included); max components $K$; inner CV folds.\\
1. Form $\mathbf{S}=\X\transpose\Y$ once.\\
2. For each operator $b$: $\mathbf{S}_b\leftarrow\A_b\mathbf{S}$; score $b$ by the
chosen criterion (CV-RMSE / PRESS / covariance) on the prefix of components it
induces.\\
3. Pick $b^\star=\arg\min_b$ criterion (one operator for all components;
identity wins if no operator helps).\\
4. Extract $K$ SIMPLS components from $\mathbf{S}_{b^\star}$; map each back with
$z_a=\A_{b^\star}\transpose r_a$.\\
5. Return $\B=\Z(\PP\transpose\Z)^{+}\Q\transpose$ and the operator log $b^\star$.\\
\textbf{Output:} one linear calibration on the original wavelength grid; predict
with $\widehat{\Y}=\X\B+\bar{\Y}$ --- no preprocessing replayed.
\end{minipage}}

\subsection{Operator-induced Ridge kernel}
\begin{propositionS}[Operator-induced kernel]\label{sp:ridge}
For centered $\X_c,\Y_c$, $\alpha>0$ and a strict linear operator $\A_b$, Ridge
in the dual depends on $\A_b$ only through $\K_b=\X_c\A_b\transpose\A_b\X_c\transpose$,
with $\mathbf{C}=(\K_b+\alpha\mathbf I_n)^{-1}\Y_c$ and original-space
coefficients $\bm\beta_b=\A_b\transpose\A_b\X_c\transpose\mathbf{C}$.
\end{propositionS}
\begin{proof}
Ridge on the transformed data $\X_b=\X_c\A_b\transpose$ has primal solution
$\bm\beta=(\X_b\transpose\X_b+\alpha\mathbf I_p)^{-1}\X_b\transpose\Y_c$. The dual
identity gives $\bm\beta=\X_b\transpose(\X_b\X_b\transpose+\alpha\mathbf I_n)^{-1}\Y_c$
with $\X_b\X_b\transpose=\X_c\A_b\transpose\A_b\X_c\transpose=\K_b$. Substituting
$\X_b\transpose=\A_b\X_c\transpose$ yields
$\bm\beta_b=\A_b\transpose\A_b\X_c\transpose\mathbf{C}$, which lives in the
original feature space.
\end{proof}
A weighted mixture replaces $\K_b$ by $\K=\sum_b s_b^2\,\X_c\A_b\transpose\A_b\X_c\transpose=\X_c\mathbf U$
with $\mathbf U=\sum_b s_b^2\,\A_b\transpose\A_b\X_c\transpose$, and the
coefficient $\bm\beta=\mathbf U\mathbf{C}$ is again original-space; the wide
block matrix is never formed. Block scales $s_b$ default to a root-mean-square
normalisation so all blocks share approximately the same Frobenius norm.

\subsection{Vertex optimum of the relaxed covariance objective}
\begin{propositionS}[Hard vertex optimum]\label{sp:soft}
Let $\A_\alpha=\sum_b\alpha_b\A_b$ with $\alpha$ on the simplex
$\Delta=\{\alpha_b\ge0,\sum_b\alpha_b=1\}$. The map
$f(\alpha)=\lVert\A_\alpha\mathbf{S}\rVert_F^2$ is convex on $\Delta$ and attains
its maximum at a vertex $\alpha=e_b$.
\end{propositionS}
\begin{proof}
$f(\alpha)=\alpha\transpose\mathbf G\alpha$ with
$\mathbf G_{bc}=\langle\A_b\mathbf{S},\A_c\mathbf{S}\rangle_F$ positive
semidefinite, so $f$ is convex. The maximum of a convex function over a compact
convex polytope is attained at an extreme point; the extreme points of $\Delta$
are the vertices $e_b$.
\end{proof}
Thus, under the pure covariance objective a soft (convex) operator mixture
reduces to a single hard operator, which is why softmax/sparsemax gates show no
advantage on this objective. This does not claim that soft mixtures collapse
under every selection criterion (e.g.\ cross-validated RMSE), nor that the
cross-validated hard policy used in the experiments is globally optimal.

\subsection{Selection variance: why a compact bank is enough}
If $B$ operators are scored by independent validation estimates
$\widehat{e}_b\sim\mathcal N(\mu_b,\sigma^2/n_{\mathrm{ho}})$, the selected
minimum is optimistically biased and, for roughly equal means, the expected
selected error behaves like
\begin{equation}
\mathbb{E}\!\left[\min_b\widehat{e}_b\right]\approx
\mu-\frac{\sigma}{\sqrt{n_{\mathrm{ho}}}}\sqrt{2\ln B},
\label{eq:s_winner}
\end{equation}
the leading extreme-value term for the minimum of $B$ Gaussians. The optimistic
bias therefore grows like $\sqrt{\ln B}$: enlarging the bank from $B=135$ to
$B=1500$ candidate evaluations inflates the bias by
$\sqrt{\ln 1500/\ln 135}\approx1.22$. Equation~\eqref{eq:s_winner} is an
extreme-value \emph{explanation}, not a sufficiency theorem (NIRS operators are
correlated and their true effects differ), but it is consistent with the
observed tie between the compact nine-operator bank and a hundred-operator bank
\citep{cawley2010overfitting,varma2006bias}.

\subsection{FastAOM: chain identity and low-rank screening}
A chain $\A_s=\A_d\cdots\A_1$ of strict linear operators composes into a single
fixed matrix, so $(\X\A_s\transpose)\transpose\Y=\A_s\X\transpose\Y$ and a large
space of chains is screened on $\mathbf{S}$. With a nonlinear base $B_j(\X)$, the
adjoint-only score
\begin{equation}
\mathrm{score}(j,s)=\frac{\lVert\A_s\transpose B_j(\X)\transpose y\rVert^2}
{\lVert B_j(\X)\A_s\transpose\rVert_F^2\,\lVert y\rVert^2}\in[0,1]
\end{equation}
follows from Cauchy--Schwarz; its denominator is approximated from a truncated
singular value decomposition $B_j(\X)\approx \mathbf U\,\mathrm{diag}(\mathbf S)\,\mathbf V\transpose$.
Surviving chains are combined by a sparse, non-negative (NNLS) weighting and fit
as a PLS-then-Ridge calibration, so the result is a sparse linear combination of
strict-linear operator chains.

\section{Numerical validation matrix}
\label{sec:s_validation}
Table~\ref{tab:s_validation} reports the agreement between the folded
computation and the explicit transformed-space reference, and between the Python
reference and the C++ \texttt{libn4m} engine, for fixed folds and fixed
strict-linear operators.

\begin{table}[t]
  \centering
  \caption{Numerical validation. Maximum observed discrepancy between equivalent
  computations of the operator-adaptive calibration.}
  \label{tab:s_validation}
  \small
  \begin{tabularx}{\linewidth}{Xrl}
  \toprule
  Equivalence check & Max.\ discrepancy & Scope \\
  \midrule
  Cross-covariance identity $(\X\A\transpose)\transpose y$ vs $\A\X\transpose y$ & $\sim10^{-12}$ & per operator (unit test) \\
  Adjoint-NIPALS vs covariance path (RMSEP) & $3.6\times10^{-11}$ & 159 dataset--seed pairs \\
  Covariance-SIMPLS vs materialised reference & $10^{-6}$ & fixed folds/operators \\
  C++ \texttt{libn4m} oracle vs Python reference (RMSEP) & $0.000$ & exact operator + component match, 4 cases \\
  \bottomrule
  \end{tabularx}
\end{table}

\section{Cohort manifest}
\label{sec:s_cohort}
The full per-dataset cohort manifest --- task, domain, sample and wavelength
counts and split provenance --- is given in the landscape longtable below.

\clearpage
\begin{landscape}
\begingroup
\small
\setlength{\tabcolsep}{3.0pt}
\renewcommand{\arraystretch}{1.08}
\setlength{\LTpre}{4pt}
\setlength{\LTpost}{4pt}
\begin{longtable}{L{6.2cm}L{2.1cm}rrrL{6.0cm}L{2.6cm}L{3.0cm}}
\caption{Full manifest overview for the AOM benchmark cohort.  This table is for provenance and includes manifest rows beyond the main $N_{\cap}=32$ score denominator; the per-dataset score table is restricted to the paired denominator stated in the text.  The longtable repeats the header after page breaks and marks rows continued on the next page.}\\
\toprule
Dataset & Task & $n$ & $p$ & $p/n$ & Response type or range & Original split & Domain \\
\midrule
\endfirsthead
\toprule
Dataset & Task & $n$ & $p$ & $p/n$ & Response type or range & Original split & Domain \\
\midrule
\endhead
\midrule
\multicolumn{8}{r}{Continued on next page} \\
\midrule
\endfoot
\bottomrule
\endlastfoot
CoffeeType\_\allowbreak{}kenstone70\_\allowbreak{}strat & classification & 70 & 601 & 8.586 & 7 classes; max share 0.14 & kenstone70\_\allowbreak{}strat & beverage \\
Species\_\allowbreak{}56\_\allowbreak{}Bagnall & classification & 56 & 286 & 5.107 & 2 classes; max share 0.52 & Bagnall & beverage \\
Oocist2C\_\allowbreak{}333\_\allowbreak{}Maia\_\allowbreak{}Acc87.6 & classification & 333 & 2151 & 6.459 & 2 classes; max share 0.52 & Maia & biomedical \\
Sporozoite2C\_\allowbreak{}229\_\allowbreak{}Maia\_\allowbreak{}Acc94.5 & classification & 229 & 2151 & 9.393 & 2 classes; max share 0.60 & Maia & biomedical \\
CT2C\_\allowbreak{}1057\_\allowbreak{}CIAT\_\allowbreak{}Acc & classification & 1056 & 1050 & 0.994 & 2 classes; max share 0.73 & CIAT & crop-\allowbreak{}tuber \\
labels\_\allowbreak{}kenstone70\_\allowbreak{}strat & classification & 450 & 601 & 1.336 & 9 classes; max share 0.11 & kenstone70\_\allowbreak{}strat & dairy \\
Strawberry2C\_\allowbreak{}983\_\allowbreak{}Holland\_\allowbreak{}Acc94.3 & classification & 983 & 235 & 0.239 & 2 classes; max share 0.64 & Holland & fruit-\allowbreak{}quality \\
Beef\_\allowbreak{}Impurity\_\allowbreak{}60\_\allowbreak{}AlJowder & classification & 60 & 470 & 7.833 & 5 classes; max share 0.20 & AlJowder & meat-\allowbreak{}quality \\
FinalScoreBin\_\allowbreak{}grp70\_\allowbreak{}30\_\allowbreak{}classStrat & classification & 935 & 2177 & 2.328 & 2 classes; max share 0.58 & grp70\_\allowbreak{}30\_\allowbreak{}classStrat & plant-\allowbreak{}disease \\
ScoreBin\_\allowbreak{}grp70\_\allowbreak{}30\_\allowbreak{}classStrat & classification & 816 & 2177 & 2.668 & 2 classes; max share 0.77 & grp70\_\allowbreak{}30\_\allowbreak{}classStrat & plant-\allowbreak{}disease \\
Genotype10\_\allowbreak{}250 & classification & 250 & 2152 & 8.608 & 10 classes; max share 0.10 & unspecified & plant-\allowbreak{}id \\
Group9\_\allowbreak{}1856 & classification & 1800 & 2152 & 1.196 & 9 classes; max share 0.16 & unspecified & plant-\allowbreak{}id \\
Group\_\allowbreak{}2185 & classification & 2185 & 2152 & 0.985 & 10 classes; max share 0.18 & unspecified & plant-\allowbreak{}id \\
InOut\_\allowbreak{}1264 & classification & 1263 & 2152 & 1.704 & 2 classes; max share 0.59 & unspecified & plant-\allowbreak{}id \\
Species\_\allowbreak{}code\_\allowbreak{}grpStrat70\_\allowbreak{}30\_\allowbreak{}bySpecimen & classification & 7323 & 1951 & 0.266 & 5 classes; max share 0.31 & grpStrat70\_\allowbreak{}30 & plant-\allowbreak{}id \\
C2\_\allowbreak{}511\_\allowbreak{}Davrieux\_\allowbreak{}Acc82 & classification & 511 & 1050 & 2.055 & 2 classes; max share 0.51 & Davrieux & wood-\allowbreak{}product \\
C5\_\allowbreak{}511\_\allowbreak{}Davrieux\_\allowbreak{}Acc82 & classification & 511 & 1050 & 2.055 & 5 classes; max share 0.42 & Davrieux & wood-\allowbreak{}product \\
Beer\_\allowbreak{}OriginalExtract\_\allowbreak{}60\_\allowbreak{}KS & regression & 60 & 576 & 9.600 & Beer\_\allowbreak{}OriginalExtract\_\allowbreak{}60; range 4.23 to 18.8 & KS & beverage \\
Beer\_\allowbreak{}OriginalExtract\_\allowbreak{}60\_\allowbreak{}YbaseSplit & regression & 60 & 576 & 9.600 & Beer\_\allowbreak{}OriginalExtract\_\allowbreak{}60; range 4.23 to 18.8 & YbaseSplit & beverage \\
Malaria\_\allowbreak{}Oocist\_\allowbreak{}333\_\allowbreak{}Maia & regression & 333 & 2151 & 6.459 & Malaria\_\allowbreak{}Oocist\_\allowbreak{}333; range 0 to 6.1e+04 & Maia & biomedical \\
Malaria\_\allowbreak{}Sporozoite\_\allowbreak{}229\_\allowbreak{}Maia & regression & 229 & 2151 & 9.393 & Malaria\_\allowbreak{}Sporozoite\_\allowbreak{}229; range 0 to 2.36e+05 & Maia & biomedical \\
Corn\_\allowbreak{}Oil\_\allowbreak{}80\_\allowbreak{}ZhengChenPelegYbaseSplit & regression & 80 & 700 & 8.750 & Corn\_\allowbreak{}Oil\_\allowbreak{}80; range 3.09 to 3.83 & YbaseSplit & crop-\allowbreak{}grain \\
Corn\_\allowbreak{}Starch\_\allowbreak{}80\_\allowbreak{}ZhengChenPelegYbaseSplit & regression & 80 & 700 & 8.750 & Corn\_\allowbreak{}Starch\_\allowbreak{}80; range 62.8 to 66.5 & YbaseSplit & crop-\allowbreak{}grain \\
Rice\_\allowbreak{}Amylose\_\allowbreak{}313\_\allowbreak{}YbasedSplit & regression & 313 & 1154 & 3.687 & Rice\_\allowbreak{}Amylose\_\allowbreak{}313; range 0 to 33.8 & YbasedSplit & crop-\allowbreak{}grain \\
C\_\allowbreak{}woOutlier & regression & 2419 & 1154 & 0.477 & C\_\allowbreak{}woOutlier; range 21.5 to 47.5 & woOutlier & crop-\allowbreak{}seed \\
N\_\allowbreak{}wOutlier & regression & 2427 & 1154 & 0.475 & N\_\allowbreak{}wOutlier; range 0.19 to 5.95 & wOutlier & crop-\allowbreak{}seed \\
N\_\allowbreak{}woOutlier & regression & 2412 & 1154 & 0.478 & N\_\allowbreak{}woOutlier; range 0.19 to 5.95 & woOutlier & crop-\allowbreak{}seed \\
Milk\_\allowbreak{}Fat\_\allowbreak{}1224\_\allowbreak{}KS & regression & 402 & 255 & 0.634 & Milk\_\allowbreak{}Fat\_\allowbreak{}1224; range 1.54 to 7.6 & KS & dairy \\
Milk\_\allowbreak{}Lactose\_\allowbreak{}1224\_\allowbreak{}KS & regression & 1224 & 255 & 0.208 & Milk\_\allowbreak{}Lactose\_\allowbreak{}1224; range 3.98 to 5.1 & KS & dairy \\
Milk\_\allowbreak{}Urea\_\allowbreak{}1224\_\allowbreak{}KS & regression & 1224 & 255 & 0.208 & Milk\_\allowbreak{}Urea\_\allowbreak{}1224; range 9 to 44 & KS & dairy \\
Biscuit\_\allowbreak{}Fat\_\allowbreak{}40\_\allowbreak{}RandomSplit & regression & 72 & 700 & 9.722 & Biscuit\_\allowbreak{}Fat\_\allowbreak{}40; range 14.8 to 21.7 & RandomSplit & food-\allowbreak{}product \\
Biscuit\_\allowbreak{}Sucrose\_\allowbreak{}40\_\allowbreak{}RandomSplit & regression & 72 & 700 & 9.722 & Biscuit\_\allowbreak{}Sucrose\_\allowbreak{}40; range 9.95 to 23.2 & RandomSplit & food-\allowbreak{}product \\
Brix\_\allowbreak{}spxy70 & regression & 50 & 600 & 12.000 & Brix; range 11.2 to 20 & spxy70 & fruit-\allowbreak{}quality \\
Firmness\_\allowbreak{}spxy70 & regression & 40 & 600 & 15.000 & Firmness; range 2.85 to 4.85 & spxy70 & fruit-\allowbreak{}quality \\
brix\_\allowbreak{}groupSampleID\_\allowbreak{}stratDateVar\_\allowbreak{}balRows & regression & 2133 & 2101 & 0.985 & brix; range 4.3 to 29 & groupSampleID\_\allowbreak{}stratDateVar & fruit-\allowbreak{}quality \\
ph\_\allowbreak{}groupSampleID\_\allowbreak{}stratDateVar\_\allowbreak{}balRows & regression & 1401 & 2101 & 1.500 & ph; range 2.29 to 4.58 & groupSampleID\_\allowbreak{}stratDateVar & fruit-\allowbreak{}quality \\
ta\_\allowbreak{}groupSampleID\_\allowbreak{}stratDateVar\_\allowbreak{}balRows & regression & 1401 & 2101 & 1.500 & ta; range 2.68 to 16.3 & groupSampleID\_\allowbreak{}stratDateVar & fruit-\allowbreak{}quality \\
TIC\_\allowbreak{}spxy70 & regression & 62 & 254 & 4.097 & TIC; range 60.2 to 97.6 & spxy70 & industrial \\
ALPINE\_\allowbreak{}P\_\allowbreak{}291\_\allowbreak{}KS & regression & 291 & 2151 & 7.392 & ALPINE\_\allowbreak{}P\_\allowbreak{}291; range -\allowbreak{}0.00724 to 0.696 & KS & leaf-\allowbreak{}physiology \\
An\_\allowbreak{}spxyG70\_\allowbreak{}30\_\allowbreak{}byCultivar\_\allowbreak{}ASD & regression & 112 & 2101 & 18.759 & An; range 0.0565 to 16.5 & spxyG\_\allowbreak{}byCultivar & leaf-\allowbreak{}physiology \\
An\_\allowbreak{}spxyG70\_\allowbreak{}30\_\allowbreak{}byCultivar\_\allowbreak{}MicroNIR & regression & 116 & 125 & 1.078 & An; range 0.0565 to 18.6 & spxyG\_\allowbreak{}byCultivar & leaf-\allowbreak{}physiology \\
An\_\allowbreak{}spxyG70\_\allowbreak{}30\_\allowbreak{}byCultivar\_\allowbreak{}MicroNIR\_\allowbreak{}NeoSpectra & regression & 115 & 276 & 2.400 & An; range 0.0565 to 18.6 & spxyG\_\allowbreak{}byCultivar & leaf-\allowbreak{}physiology \\
An\_\allowbreak{}spxyG70\_\allowbreak{}30\_\allowbreak{}byCultivar\_\allowbreak{}NeoSpectra & regression & 119 & 257 & 2.160 & An; range 0.0565 to 18.6 & spxyG\_\allowbreak{}byCultivar & leaf-\allowbreak{}physiology \\
Ccar\_\allowbreak{}spxyG\_\allowbreak{}block2deg & regression & 4245 & 196 & 0.046 & Ccar; range -\allowbreak{}999 to 28.4 & spxyG\_\allowbreak{}block2deg & leaf-\allowbreak{}physiology \\
Chla+b\_\allowbreak{}spxyG\_\allowbreak{}block2deg & regression & 6850 & 196 & 0.029 & Chla+b; range -\allowbreak{}999 to 167 & spxyG\_\allowbreak{}block2deg & leaf-\allowbreak{}physiology \\
Chla+b\_\allowbreak{}spxyG\_\allowbreak{}species & regression & 6850 & 196 & 0.029 & Chla+b; range -\allowbreak{}999 to 167 & spxyG & leaf-\allowbreak{}physiology \\
LMA\_\allowbreak{}spxyG70\_\allowbreak{}30\_\allowbreak{}byCultivar\_\allowbreak{}ASD & regression & 1564 & 2101 & 1.343 & LMA; range 1.43 to 7.59 & spxyG\_\allowbreak{}byCultivar & leaf-\allowbreak{}physiology \\
LMA\_\allowbreak{}spxyG\_\allowbreak{}block2deg & regression & 45417 & 196 & 0.004 & LMA; range 0.066 to 389 & spxyG\_\allowbreak{}block2deg & leaf-\allowbreak{}physiology \\
LP\_\allowbreak{}spxyG & regression & 257 & 2101 & 8.175 & LP; range 0.0681 to 0.671 & spxyG & leaf-\allowbreak{}physiology \\
MP\_\allowbreak{}spxyG & regression & 257 & 2101 & 8.175 & MP; range 0.0202 to 0.114 & spxyG & leaf-\allowbreak{}physiology \\
NP\_\allowbreak{}spxyG & regression & 257 & 2101 & 8.175 & NP; range 0.0561 to 0.57 & spxyG & leaf-\allowbreak{}physiology \\
Pi\_\allowbreak{}spxyG & regression & 257 & 2101 & 8.175 & Pi; range 0.0225 to 0.63 & spxyG & leaf-\allowbreak{}physiology \\
Rd25\_\allowbreak{}CBtestSite & regression & 470 & 2151 & 4.577 & Rd25\_\allowbreak{}CBtestSite; range 0.465 to 2.03 & external\_\allowbreak{}site\_\allowbreak{}CB & leaf-\allowbreak{}physiology \\
Rd25\_\allowbreak{}GTtestSite & regression & 470 & 2151 & 4.577 & Rd25\_\allowbreak{}GTtestSite; range 0.465 to 2.03 & external\_\allowbreak{}site\_\allowbreak{}GT & leaf-\allowbreak{}physiology \\
Rd25\_\allowbreak{}XSBNtestSite & regression & 470 & 2151 & 4.577 & Rd25\_\allowbreak{}XSBNtestSite; range 0.465 to 2.03 & external\_\allowbreak{}site\_\allowbreak{}XSBN & leaf-\allowbreak{}physiology \\
Rd25\_\allowbreak{}spxy70 & regression & 470 & 2151 & 4.577 & Rd25; range 0.465 to 2.03 & spxy70 & leaf-\allowbreak{}physiology \\
V25\_\allowbreak{}spxyG & regression & 250 & 2101 & 8.404 & V25; range 0.0229 to 1.39 & spxyG & leaf-\allowbreak{}physiology \\
WUEinst\_\allowbreak{}spxyG70\_\allowbreak{}30\_\allowbreak{}byCultivar\_\allowbreak{}MicroNIR\_\allowbreak{}NeoSpectra & regression & 112 & 276 & 2.464 & WUEinst; range 0.833 to 12.4 & spxyG\_\allowbreak{}byCultivar & leaf-\allowbreak{}physiology \\
grapevine\_\allowbreak{}chloride\_\allowbreak{}556\_\allowbreak{}KS & regression & 555 & 1023 & 1.843 & grapevine\_\allowbreak{}chloride\_\allowbreak{}556; range 0 to 7.95e+03 & KS & leaf-\allowbreak{}physiology \\
Beef\_\allowbreak{}Marbling\_\allowbreak{}RandomSplit & regression & 832 & 331 & 0.398 & Beef\_\allowbreak{}Marbling; range 100 to 810 & RandomSplit & meat-\allowbreak{}quality \\
Quartz\_\allowbreak{}spxy70 & regression & 303 & 1500 & 4.950 & Quartz & spxy70 & mineral \\
DIESEL\_\allowbreak{}bp50\_\allowbreak{}246\_\allowbreak{}b-\allowbreak{}a & regression & 226 & 401 & 1.774 & DIESEL\_\allowbreak{}bp50\_\allowbreak{}246\_\allowbreak{}b-\allowbreak{}a; range 197 to 293 & unspecified & petroleum \\
DIESEL\_\allowbreak{}bp50\_\allowbreak{}246\_\allowbreak{}hla-\allowbreak{}b & regression & 246 & 401 & 1.630 & DIESEL\_\allowbreak{}bp50\_\allowbreak{}246\_\allowbreak{}hla-\allowbreak{}b; range 197 to 293 & unspecified & petroleum \\
DIESEL\_\allowbreak{}bp50\_\allowbreak{}246\_\allowbreak{}hlb-\allowbreak{}a & regression & 246 & 401 & 1.630 & DIESEL\_\allowbreak{}bp50\_\allowbreak{}246\_\allowbreak{}hlb-\allowbreak{}a; range 197 to 293 & unspecified & petroleum \\
Escitalopramt\_\allowbreak{}310\_\allowbreak{}Zhao & regression & 310 & 404 & 1.303 & Escitalopramt\_\allowbreak{}310; range 4.61 to 9.79 & Zhao & pharmaceutical \\
FinalScore\_\allowbreak{}grp70\_\allowbreak{}30\_\allowbreak{}scoreQ & regression & 935 & 2177 & 2.328 & FinalScore; range 1 to 5 & grp70\_\allowbreak{}30\_\allowbreak{}scoreQ & plant-\allowbreak{}disease \\
Fv\_\allowbreak{}Fm\_\allowbreak{}grp70\_\allowbreak{}30 & regression & 518 & 2177 & 4.203 & Fv\_\allowbreak{}Fm; range 0.512 to 0.856 & grp70\_\allowbreak{}30 & plant-\allowbreak{}disease \\
Tleaf\_\allowbreak{}grp70\_\allowbreak{}30 & regression & 665 & 2177 & 3.274 & Tleaf; range 19.7 to 44.7 & grp70\_\allowbreak{}30 & plant-\allowbreak{}disease \\
All\_\allowbreak{}manure\_\allowbreak{}CaO\_\allowbreak{}SPXY\_\allowbreak{}strat\_\allowbreak{}Manure\_\allowbreak{}type & regression & 490 & 1003 & 2.047 & All\_\allowbreak{}manure\_\allowbreak{}CaO; range 1.53 to 146 & SPXY\_\allowbreak{}strat & soil-\allowbreak{}amendment \\
All\_\allowbreak{}manure\_\allowbreak{}K2O\_\allowbreak{}SPXY\_\allowbreak{}strat\_\allowbreak{}Manure\_\allowbreak{}type & regression & 490 & 1003 & 2.047 & All\_\allowbreak{}manure\_\allowbreak{}K2O; range 0.66 to 47.7 & SPXY\_\allowbreak{}strat & soil-\allowbreak{}amendment \\
All\_\allowbreak{}manure\_\allowbreak{}MgO\_\allowbreak{}SPXY\_\allowbreak{}strat\_\allowbreak{}Manure\_\allowbreak{}type & regression & 490 & 1003 & 2.047 & All\_\allowbreak{}manure\_\allowbreak{}MgO; range 0.52 to 18.8 & SPXY\_\allowbreak{}strat & soil-\allowbreak{}amendment \\
All\_\allowbreak{}manure\_\allowbreak{}P2O5\_\allowbreak{}SPXY\_\allowbreak{}strat\_\allowbreak{}Manure\_\allowbreak{}type & regression & 490 & 1003 & 2.047 & All\_\allowbreak{}manure\_\allowbreak{}P2O5; range 0.84 to 43.7 & SPXY\_\allowbreak{}strat & soil-\allowbreak{}amendment \\
All\_\allowbreak{}manure\_\allowbreak{}Total\_\allowbreak{}N\_\allowbreak{}SPXY\_\allowbreak{}strat\_\allowbreak{}Manure\_\allowbreak{}type & regression & 490 & 1003 & 2.047 & All\_\allowbreak{}manure\_\allowbreak{}Total\_\allowbreak{}N; range 2.07 to 40.5 & SPXY\_\allowbreak{}strat & soil-\allowbreak{}amendment \\
LUCAS\_\allowbreak{}SOC\_\allowbreak{}Cropland\_\allowbreak{}8731\_\allowbreak{}NocitaKS & regression & 8731 & 4200 & 0.481 & LUCAS\_\allowbreak{}SOC\_\allowbreak{}Cropland\_\allowbreak{}8731; range 0 to 194 & NocitaKS & soil-\allowbreak{}eu \\
LUCAS\_\allowbreak{}SOC\_\allowbreak{}all\_\allowbreak{}26650\_\allowbreak{}NocitaKS & regression & 19036 & 4200 & 0.221 & LUCAS\_\allowbreak{}SOC\_\allowbreak{}all\_\allowbreak{}26650; range 0 to 587 & NocitaKS & soil-\allowbreak{}eu \\
LUCAS\_\allowbreak{}pH\_\allowbreak{}Organic\_\allowbreak{}1763\_\allowbreak{}LiuRandomOrganic & regression & 1763 & 4200 & 2.382 & LUCAS\_\allowbreak{}pH\_\allowbreak{}Organic\_\allowbreak{}1763; range 2.57 to 7.28 & LiuRandom & soil-\allowbreak{}eu \\
WOOD\_\allowbreak{}Density\_\allowbreak{}402\_\allowbreak{}Olale & regression & 402 & 1038 & 2.582 & WOOD\_\allowbreak{}Density\_\allowbreak{}402; range 0.197 to 0.952 & Olale & wood-\allowbreak{}product \\
WOOD\_\allowbreak{}N\_\allowbreak{}402\_\allowbreak{}Olale & regression & 402 & 1038 & 2.582 & WOOD\_\allowbreak{}N\_\allowbreak{}402; range 0.08 to 0.49 & Olale & wood-\allowbreak{}product \\
\end{longtable}
\endgroup
\end{landscape}
\clearpage

\section{Operator-bank diagnostics}
\label{sec:s_operators}
\begin{figure}[t]
  \centering
  \includegraphics[width=0.95\linewidth]{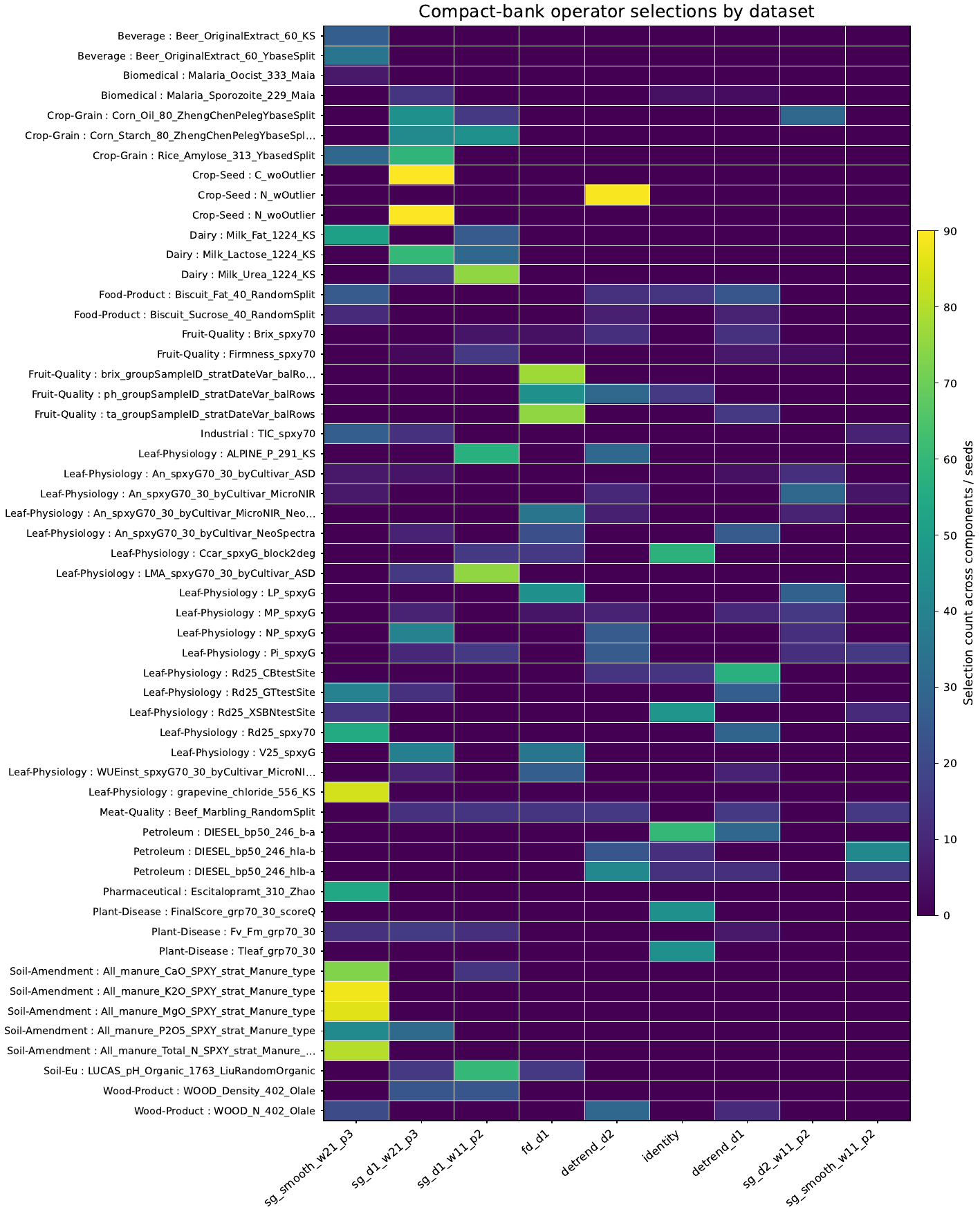}
  \caption{Per-dataset compact-bank operator-selection heatmap (counts across
  components and seeds).}
  \label{fig:s_opheat}
\end{figure}

Table~\ref{tab:s_opdiag} gives the selection frequency of each strict-linear
operator across the cohort, and Figure~\ref{fig:s_opheat} the per-dataset
selection heatmap. Savitzky--Golay smoothing and first derivatives dominate; the
identity is selected on a minority of components, consistent with a bank that is
diverse enough to adapt but small enough to keep selection stable
(Section~\ref{sec:s_proofs}, Eq.~\eqref{eq:s_winner}).

\begin{table}[t]
  \centering
  \caption{Operator-selection frequency (compact bank).}
  \label{tab:s_opdiag}
  \small
  \begin{tabularx}{\linewidth}{Xrrr}
\toprule
Compact-bank operator & Selections & Component fraction & Datasets using it \\
\midrule
sg\_\allowbreak{}smooth\_\allowbreak{}w21\_\allowbreak{}p3 & 876 & 23.4\% & 22 \\
sg\_\allowbreak{}d1\_\allowbreak{}w21\_\allowbreak{}p3 & 679 & 18.1\% & 24 \\
sg\_\allowbreak{}d1\_\allowbreak{}w11\_\allowbreak{}p2 & 497 & 13.3\% & 16 \\
fd\_\allowbreak{}d1 & 415 & 11.1\% & 14 \\
detrend\_\allowbreak{}d2 & 386 & 10.3\% & 16 \\
identity & 327 & 8.7\% & 11 \\
detrend\_\allowbreak{}d1 & 306 & 8.2\% & 18 \\
sg\_\allowbreak{}d2\_\allowbreak{}w11\_\allowbreak{}p2 & 151 & 4.0\% & 9 \\
sg\_\allowbreak{}smooth\_\allowbreak{}w11\_\allowbreak{}p2 & 112 & 3.0\% & 7 \\
\bottomrule
\end{tabularx}

\end{table}

\section{Missing-dataset reason-code audit}
\label{sec:s_missing}
Table~\ref{tab:s_missing} reports, per variant, how the manifest rows reduce to
the strict paired intersection $N_{\cap}=32$, with a reason code for each missing
row.

\begin{table}[t]
  \centering
  \caption{Missing-dataset audit by variant (reason codes: NaN spectra, fit
  error, not attempted).}
  \label{tab:s_missing}
  \small
  \begin{tabularx}{\linewidth}{Xrrrrl}
\toprule
Variant (paper label / runner key) & Complete & Error & Not attempted & Total available & Dominant reason for absence \\
\midrule
PLS-\allowbreak{}default (\code{pls-default-cv5}) & 57 & 3 & 1 & 60 & NaN in source data \\
Ridge-\allowbreak{}default (\code{ridge-default-cv5}) & 58 & 2 & 1 & 60 & NaN in source data \\
PLS-\allowbreak{}HPO (\code{pls-hpo-25trials}) & 36 & 2 & 23 & 38 & Compute budget (not attempted) \\
Ridge-\allowbreak{}HPO (\code{ridge-hpo-60trials}) & 35 & 2 & 24 & 37 & Compute budget (not attempted) \\
AOM-\allowbreak{}PLS (simple, \code{AOM-compact-cv5}) & 55 & 0 & 6 & 55 & Workspace scope (not attempted) \\
AOM-\allowbreak{}PLS (best, \code{ASLS-AOM-compact-cv5}) & 53 & 2 & 6 & 55 & ASLS divergence / not attempted \\
AOM-\allowbreak{}Ridge (simple, \code{AOMRidge-global-compact-none}) & 53 & 0 & 8 & 53 & Workspace scope (not attempted) \\
AOM-\allowbreak{}Ridge (best, \code{AOMRidge-Blender-headline-spxy3}) & 53 & 0 & 8 & 53 & Workspace scope (not attempted) \\
\bottomrule
\end{tabularx}

\end{table}

Table~\ref{tab:s_representativity} compares the strict paired subset against the
full regression manifest. The strict subset remains diverse, but it is not a
complete mirror of the 61-row inventory: it covers 15 of 25 source families and
10 of 17 regression domains, with smaller median training and test sizes. These
differences are why the main text treats $N_{\cap}=32$ as a conservative
protocol-consistent denominator rather than as the full empirical scope of the
software.

\begin{table}[t]
  \centering
  \caption{Representativity of the strict paired regression denominator.}
  \label{tab:s_representativity}
  \small
  \begin{tabularx}{\linewidth}{Xrr}
\toprule
Property & Full regression manifest & Strict paired subset \\
\midrule
Datasets & 61 & 32 \\
Source families & 25 & 15 \\
Analytical domains & 17 & 10 \\
Split types & 23 & 15 \\
Train samples, median (range) & 247 (28--39{,}225) & 225 (40--1{,}434) \\
Test samples, median (range) & 147 (12--6{,}192) & 113 (16--3{,}229) \\
Wavelengths, median (range) & 1{,}023 (125--4{,}200) & 862 (125--2{,}177) \\
$p/n_{\mathrm{train}}$, median (range) & 3.55 (0.005--26.94) & 3.57 (0.19--26.94) \\
\bottomrule
\end{tabularx}

\end{table}

The narrow full-HPO intersection reflects the coverage of one expensive protocol,
not a selective denominator. Table~\ref{tab:s_hpocov} unions the three
tuned-linear protocols already available: full-HPO (preprocessing search),
model-only default-CV5, and the externally-tuned PLS/Ridge baselines. Tuned-linear
coverage reaches 59 of 61 regression datasets for both PLS and Ridge; the only two
datasets with no tuned result under any protocol are the FUSARIUM targets that
fail every linear method with \code{ValueError: Input X contains NaN}. The
headline paired tests retain the strict full-HPO intersection for protocol
consistency.

\begin{table}[t]
  \centering
  \caption{Tuned-linear coverage of the 61 regression datasets, per protocol and
  unioned. The strict single-protocol full-HPO intersection is $N_{\cap}=35$;
  the union of the three tuned protocols already on disk reaches $N_{\cap}=59$.}
  \label{tab:s_hpocov}
  \small
  \begin{tabularx}{\linewidth}{Xrr}
\toprule
Tuned-linear protocol & PLS datasets & Ridge datasets \\
\midrule
\multicolumn{3}{@{}l}{\emph{Per-protocol coverage (of 61 regression datasets)}} \\
\quad PLS/Ridge-HPO (full preprocessing search) & 36 & 35 \\
\quad PLS/Ridge-default (model hyperparameter, cv5) & 57 & 58 \\
\quad paper-PLS/paper-Ridge (literature-tuned) & 54 & 54 \\
\midrule
\multicolumn{3}{@{}l}{\emph{Union over the three tuned protocols (any-tuned)}} \\
\quad Any-tuned coverage & 59 & 59 \\
\quad $N_{\cap}$ (both PLS and Ridge tuned) & 59 & 59 \\
\bottomrule
\multicolumn{3}{@{}p{\linewidth}}{\footnotesize The strict single-protocol full-HPO intersection is $N_{\cap}=35$; unioning the three tuned protocols already on disk raises it to $N_{\cap}=59$. The 2 datasets with no tuned PLS or Ridge result under any protocol are FinalScore\_\allowbreak{}grp70\_\allowbreak{}30\_\allowbreak{}scoreQ, Tleaf\_\allowbreak{}grp70\_\allowbreak{}30\,---\,both FUSARIUM targets that fail every linear protocol with \code{ValueError: Input X contains NaN}.} \\
\end{tabularx}

\end{table}

The conventional preprocessing search does not converge on a single recipe.
Table~\ref{tab:s_recipe} tabulates how often each normalisation, smoother and
baseline was selected across the HPO fits: the most frequent choice is SNV with
detrending for PLS and SNV with a Gaussian smoother for Ridge, but no combined
recipe exceeds 10\% of fits, and the selection is dataset-dependent. The de-facto
``fixed recipe'' is therefore weak --- which is precisely the regime
operator-adaptive selection is designed for.

\begin{table}[t]
  \centering
  \caption{Most-frequently-selected conventional preprocessing under HPO (the
  de-facto fixed recipe), from \code{best\_config\_json} across seeds 0/1/2.}
  \label{tab:s_recipe}
  \small
  \begin{tabularx}{\linewidth}{l X}
\toprule
Component & Most-frequently-selected choices (count, share of fits) \\
\midrule
\multicolumn{2}{l}{\emph{PLS-HPO} -- 108 fits over 36 datasets (seeds 0/1/2)} \\
Normalisation & none (49, 45\%); snv (48, 44\%); msc (8, 7\%) \\
Smoothing / derivative & none (29, 27\%); Gaussian d0 sigma2 (19, 18\%); SG w31 p2 d1 (16, 15\%) \\
Baseline & none (43, 40\%); detrend (37, 34\%); asls (28, 26\%) \\
OSC components & osc\_\allowbreak{}3 (50, 46\%); none (24, 22\%); osc\_\allowbreak{}2 (20, 19\%) \\
Top recipe (norm $|$ smooth $|$ baseline) & snv $|$ none $|$ detrend (7, 6\%); none $|$ none $|$ detrend (6, 6\%) \\
\midrule
\multicolumn{2}{l}{\emph{Ridge-HPO} -- 105 fits over 35 datasets (seeds 0/1/2)} \\
Normalisation & snv (64, 61\%); none (31, 30\%); msc (10, 10\%) \\
Smoothing / derivative & Gaussian d0 sigma2 (31, 30\%); none (29, 28\%); Gaussian d0 sigma1 (12, 11\%) \\
Baseline & detrend (43, 41\%); none (39, 37\%); asls (23, 22\%) \\
OSC components & osc\_\allowbreak{}3 (39, 37\%); none (30, 29\%); osc\_\allowbreak{}1 (19, 18\%) \\
Top recipe (norm $|$ smooth $|$ baseline) & snv $|$ Gaussian d0 sigma2 $|$ none (9, 9\%); snv $|$ Gaussian d0 sigma2 $|$ detrend (9, 9\%) \\
\bottomrule
\end{tabularx}

\end{table}

The de-facto recipe above is read off the HPO selections. As a stricter check we
also applied a \emph{literally fixed} conventional recipe uniformly to every
cohort dataset --- per-sample SNV followed by a first-derivative Savitzky--Golay
filter (window 15, order 2) --- choosing only the PLS component count (resp.\ the
Ridge penalty $\alpha$) by five-fold cross-validation. This is the single default
recipe a practitioner reaches for. Table~\ref{tab:s_fixedrecipe} reports the
result on the paired denominator: the fixed recipe gives \emph{no} systematic
gain over plain PLS or Ridge (median RMSEP ratios $1.003$ and $1.017$), confirming
that no single preprocessing is broadly optimal. Selecting the operator per
dataset inside the fit improves on the same fixed recipe by $2.0\%$ (AOM-PLS,
$31/53$ datasets) and $5.1\%$ (AOM-Ridge, $34/52$) in median RMSEP. Two datasets
(\code{FinalScore\_grp70\_30\_scoreQ}, \code{Tleaf\_grp70\_30}) are omitted for
non-finite spectra.

\begin{table}[t]
  \centering
  \caption{Literally fixed conventional recipe (SNV $+$ first-derivative
  Savitzky--Golay, window 15 / order 2; $n_{\text{components}}$ and $\alpha$ by
  five-fold CV) applied uniformly across the cohort. Median paired RMSEP ratios
  (below one favours the row method) and per-dataset wins.}
  \label{tab:s_fixedrecipe}
  \small
  \begin{tabularx}{\linewidth}{Xrrr}
\toprule
Comparison & $N$ & Median RMSEP ratio & Wins \\
\midrule
PLS-\allowbreak{}fixed-\allowbreak{}recipe vs PLS-\allowbreak{}standard & 53 & 1.003 & 25/53 \\
AOM-\allowbreak{}PLS (compact-\allowbreak{}cv5) vs PLS-\allowbreak{}fixed-\allowbreak{}recipe & 53 & 0.980 & 31/53 \\
Ridge-\allowbreak{}fixed-\allowbreak{}recipe vs Ridge-\allowbreak{}raw & 52 & 1.017 & 22/52 \\
AOM-\allowbreak{}Ridge (global) vs Ridge-\allowbreak{}fixed-\allowbreak{}recipe & 52 & 0.949 & 34/52 \\
\bottomrule
\end{tabularx}

\end{table}

\section{Pairwise-denominator sensitivity}
\label{sec:s_pairwise}
The main manuscript keeps a strict common denominator for the eight displayed
paper variants. Table~\ref{tab:s_pairwise_largest} instead uses the largest
available paired denominator for each comparison. The direction of the default
baseline comparisons is stable on the wider denominators: AOM-PLS remains near
parity with default PLS, and AOM-Ridge remains better than default Ridge. The
comparisons against full HPO remain constrained by the HPO coverage itself
($N=32$--34), so they should be read as paired evidence on the expensive
protocol, not as evidence over all 61 regression rows.

\begin{table}[t]
  \centering
  \caption{Largest available paired denominator by comparison. Ratios below one
  favour the row method.}
  \label{tab:s_pairwise_largest}
  \small
  \begin{tabularx}{\linewidth}{Xrrr}
\toprule
Comparison & Largest paired $N$ & Median ratio & Wins \\
\midrule
AOM-PLS simple vs PLS-default & 52 & 0.996 & 32/52 \\
AOM-PLS best vs PLS-default & 52 & 0.985 & 33/52 \\
AOM-PLS simple vs PLS-HPO & 32 & 0.990 & 19/32 \\
AOM-PLS best vs PLS-HPO & 32 & 1.002 & 15/32 \\
AOM-Ridge simple vs Ridge-default & 52 & 0.974 & 41/52 \\
AOM-Ridge best vs Ridge-default & 52 & 0.913 & 44/52 \\
AOM-Ridge simple vs Ridge-HPO & 34 & 0.978 & 21/34 \\
AOM-Ridge best vs Ridge-HPO & 34 & 0.956 & 27/34 \\
PLS-HPO vs PLS-default & 36 & 0.993 & 21/36 \\
Ridge-HPO vs Ridge-default & 35 & 0.982 & 20/35 \\
\bottomrule
\end{tabularx}

\end{table}

\section{Variant families and negative ablations}
\label{sec:s_variants}
\begin{figure}[t]
  \centering
  \includegraphics[width=0.95\linewidth]{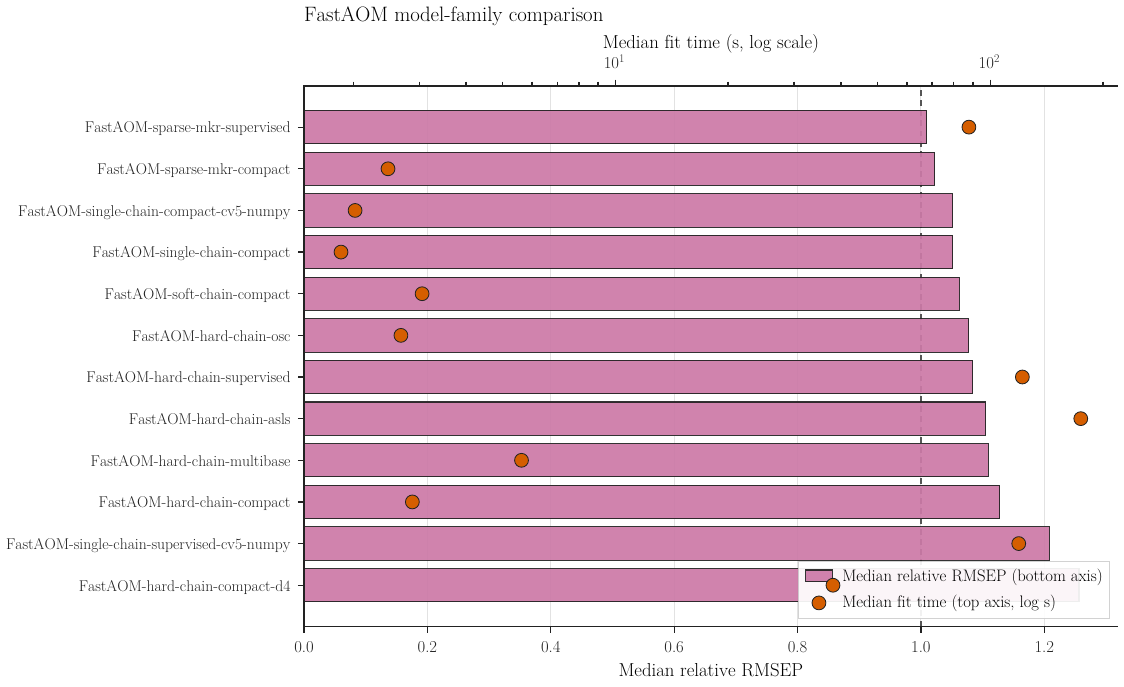}
  \caption{FastAOM variants ordered by median relative RMSEP; the heavy chain
  policies cost two orders of magnitude more time for no accuracy gain.}
  \label{fig:s_fastaom}
\end{figure}

Tables~\ref{tab:s_aompls}--\ref{tab:s_fastaom} report the full AOM-PLS, AOM-Ridge
and FastAOM families on their stated denominators, including the negative
ablations kept for honesty: per-component selection (POP-PLS) underperforms the
global choice (median ratio $1.37$--$1.39$), and a local-neighbourhood Ridge
variant is retained as a ``does not always win'' example. The FastAOM finalists
are reported as sparse non-negative combinations of strict-linear operator chains
(Fig.~\ref{fig:s_fastaom}).

\begin{table}[t]
  \centering
  \caption{AOM-PLS family.}
  \label{tab:s_aompls}
  \small
  \begin{tabularx}{\linewidth}{Xrrrrr}
\toprule
Variant & Datasets & Runs & Median RMSEP & Median ratio & Wins vs PLS \\
\midrule
ASLS-AOM-compact-cv5-numpy & 53 & 159 & 0.466 & 0.962 & 107/159 \\
AOM-compact-cv3-numpy & 55 & 165 & 0.442 & 0.980 & 98/165 \\
AOM-compact-cv5-numpy & 55 & 165 & 0.442 & 0.981 & 107/165 \\
AOM-compact-simpls-covariance-numpy & 55 & 165 & 0.795 & 0.999 & 84/165 \\
AOM-default-nipals-adjoint-numpy & 55 & 165 & 0.549 & 0.999 & 81/165 \\
aom\_nirs-AOM-PLS-default & 55 & 165 & 0.549 & 0.999 & 81/165 \\
PLS-standard-numpy & 55 & 165 & 0.795 & 1.000 & 0/165 \\
POP-nipals-adjoint-numpy & 55 & 165 & 1.784 & 1.373 & 34/165 \\
POP-simpls-covariance-numpy & 55 & 165 & 1.805 & 1.385 & 33/165 \\
\bottomrule
\end{tabularx}

\end{table}

\begin{table}[t]
  \centering
  \caption{AOM-Ridge family.}
  \label{tab:s_aomridge}
  \small
  \begin{tabularx}{\linewidth}{L{0.13\linewidth}Xrrrr}
\toprule
Source & Variant & Datasets & Runs & Median RMSEP & Median fit (s) \\
\midrule
headline & AOMRidge-global-compact-none & 53 & 53 & 0.359 & 21.60 \\
headline & AOMRidge-global-compact-none-asls & 53 & 53 & 0.382 & 24.35 \\
headline & AOMRidge-global-compact-none-snv & 53 & 53 & 0.389 & 18.23 \\
headline & AOM-PLS-compact-CV & 53 & 53 & 0.414 & 1.94 \\
headline & AOMRidgePLS-compact-colscale-cv-relative & 53 & 53 & 0.414 & 33.41 \\
headline & AOMRidge-Blender-headline-spxy3 & 53 & 55 & 0.471 & 960.1 \\
headline & AOMRidge-global-compact-none-msc & 53 & 53 & 0.483 & 25.90 \\
headline & AOMRidge-AutoSelect-headline-spxy3 & 53 & 55 & 0.520 & 646.2 \\
headline & Ridge-raw & 53 & 53 & 0.537 & 1.01 \\
headline & AOMRidgePLS-compact-Hmax-relative-emsc2 & 53 & 53 & 1.981 & 29.96 \\
seeds012 & AOMRidge-global-compact-none-split\_aware & 25 & 75 & 0.371 & 24.68 \\
seeds012 & AOMRidge-Local-compact-cv-blended & 25 & 75 & 0.484 & 4.19 \\
seeds012 & AOMRidge-Local-compact-knn50 & 25 & 75 & 0.523 & 3.19 \\
seeds012 & Ridge-raw & 25 & 76 & 0.571 & 0.49 \\
\bottomrule
\end{tabularx}

\end{table}

\begin{table}[t]
  \centering
  \caption{POP per-component selection summary (negative ablation).}
  \label{tab:s_pop}
  \small
  \begin{tabularx}{\linewidth}{Xrrrr}
\toprule
POP variant & Datasets & Runs & Median ratio vs PLS & Wins vs PLS \\
\midrule
POP-nipals-adjoint-numpy & 55 & 165 & 1.373 & 34/165 \\
POP-simpls-covariance-numpy & 55 & 165 & 1.385 & 33/165 \\
\bottomrule
\end{tabularx}

\end{table}

\begin{table}[t]
  \centering
  \caption{FastAOM variants (the best model is a sparse linear-chain combination).}
  \label{tab:s_fastaom}
  \small
  \begin{tabularx}{\linewidth}{p{0.38\linewidth}lrrrr}
\toprule
Variant & Family & $N$ & Median rel. RMSEP & Median fit (s) & Wins \\
\midrule
FastAOM-\allowbreak{}sparse-\allowbreak{}chains-\allowbreak{}supervised & Sparse chains & 50 & 1.009 & 87.77 & 10 \\
ASLS-\allowbreak{}AOM-\allowbreak{}compact-\allowbreak{}cv5-\allowbreak{}numpy & Reference & 57 & 1.011 & 1.36 & 8 \\
AOM-\allowbreak{}compact-\allowbreak{}cv5-\allowbreak{}numpy & Reference & 59 & 1.021 & 1.31 & 2 \\
FastAOM-\allowbreak{}sparse-\allowbreak{}chains-\allowbreak{}compact & Sparse chains & 50 & 1.022 & 2.48 & 3 \\
nirs4all-\allowbreak{}AOM-\allowbreak{}PLS-\allowbreak{}default & Reference & 59 & 1.034 & 0.97 & 7 \\
FastAOM-\allowbreak{}single-\allowbreak{}chain-\allowbreak{}compact & Single chain & 52 & 1.052 & 1.86 & 2 \\
FastAOM-\allowbreak{}single-\allowbreak{}chain-\allowbreak{}compact-\allowbreak{}cv5-\allowbreak{}numpy & Single chain & 52 & 1.052 & 2.02 & 0 \\
PLS-\allowbreak{}standard-\allowbreak{}numpy & Reference & 59 & 1.053 & 0.03 & 2 \\
FastAOM-\allowbreak{}soft-\allowbreak{}chain-\allowbreak{}compact & Soft chain & 52 & 1.062 & 3.05 & 1 \\
FastAOM-\allowbreak{}hard-\allowbreak{}chain-\allowbreak{}osc & Hard chain & 52 & 1.078 & 2.68 & 0 \\
FastAOM-\allowbreak{}hard-\allowbreak{}chain-\allowbreak{}supervised & Hard chain & 52 & 1.084 & 121.87 & 4 \\
FastAOM-\allowbreak{}hard-\allowbreak{}chain-\allowbreak{}asls & Hard chain & 52 & 1.105 & 174.46 & 3 \\
FastAOM-\allowbreak{}hard-\allowbreak{}chain-\allowbreak{}multibase & Hard chain & 52 & 1.109 & 5.62 & 0 \\
FastAOM-\allowbreak{}hard-\allowbreak{}chain-\allowbreak{}compact & Hard chain & 52 & 1.128 & 2.88 & 3 \\
FastAOM-\allowbreak{}single-\allowbreak{}chain-\allowbreak{}supervised-\allowbreak{}cv5-\allowbreak{}numpy & Single chain & 52 & 1.208 & 119.19 & 2 \\
FastAOM-\allowbreak{}hard-\allowbreak{}chain-\allowbreak{}compact-\allowbreak{}d4 & Hard chain & 1 & 1.256 & 38.08 & 0 \\
\bottomrule
\end{tabularx}

\end{table}

\section{Per-dataset results}
\label{sec:s_perdataset}
\begin{figure}[t]
  \centering
  \includegraphics[width=\linewidth]{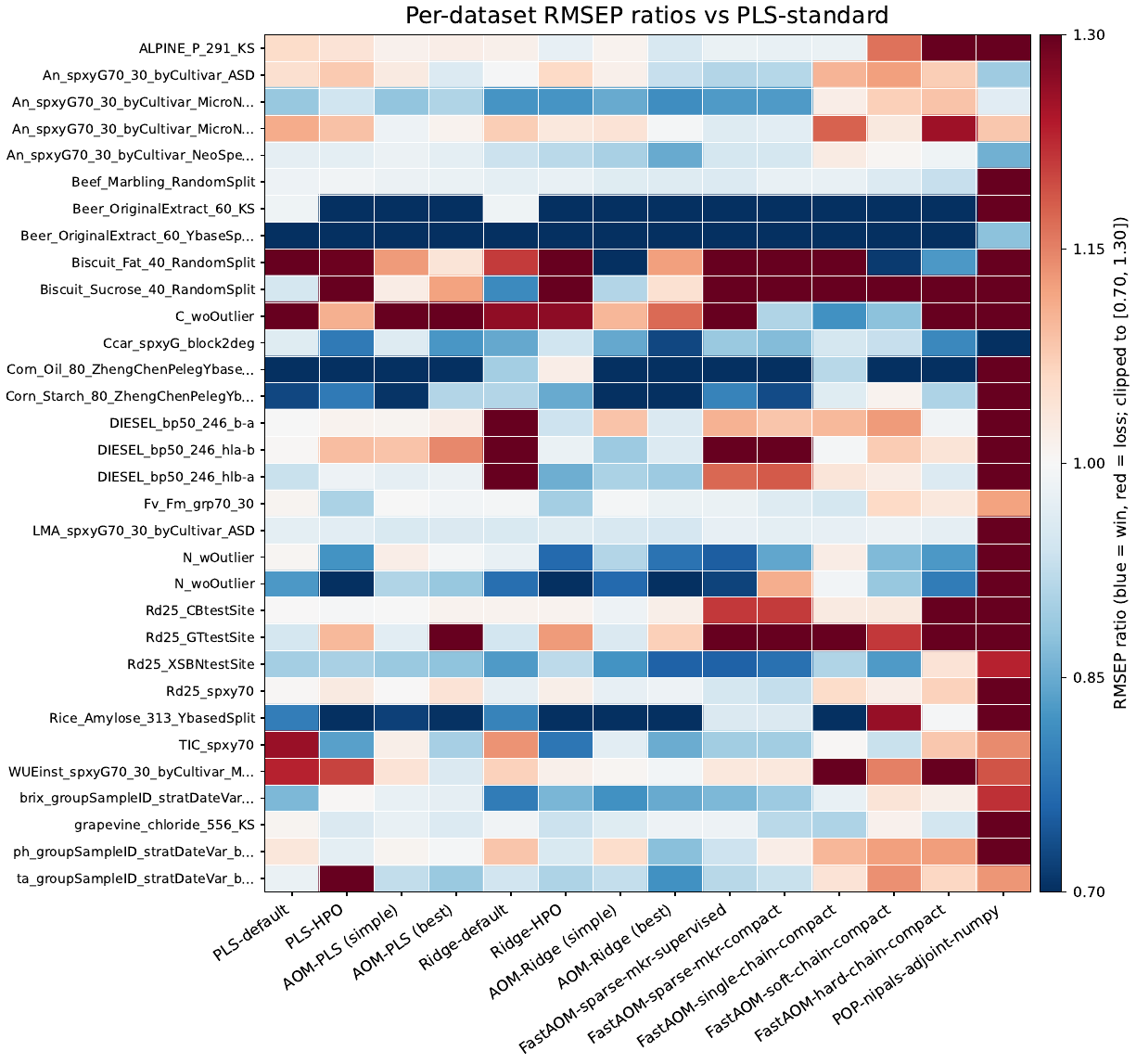}
  \caption{Per-dataset RMSEP ratios versus PLS-standard (diverging scale centered
  at 1.0); missing cells in grey.}
  \label{fig:s_heat}
\end{figure}

\begin{figure}[t]
  \centering
  \includegraphics[width=\linewidth]{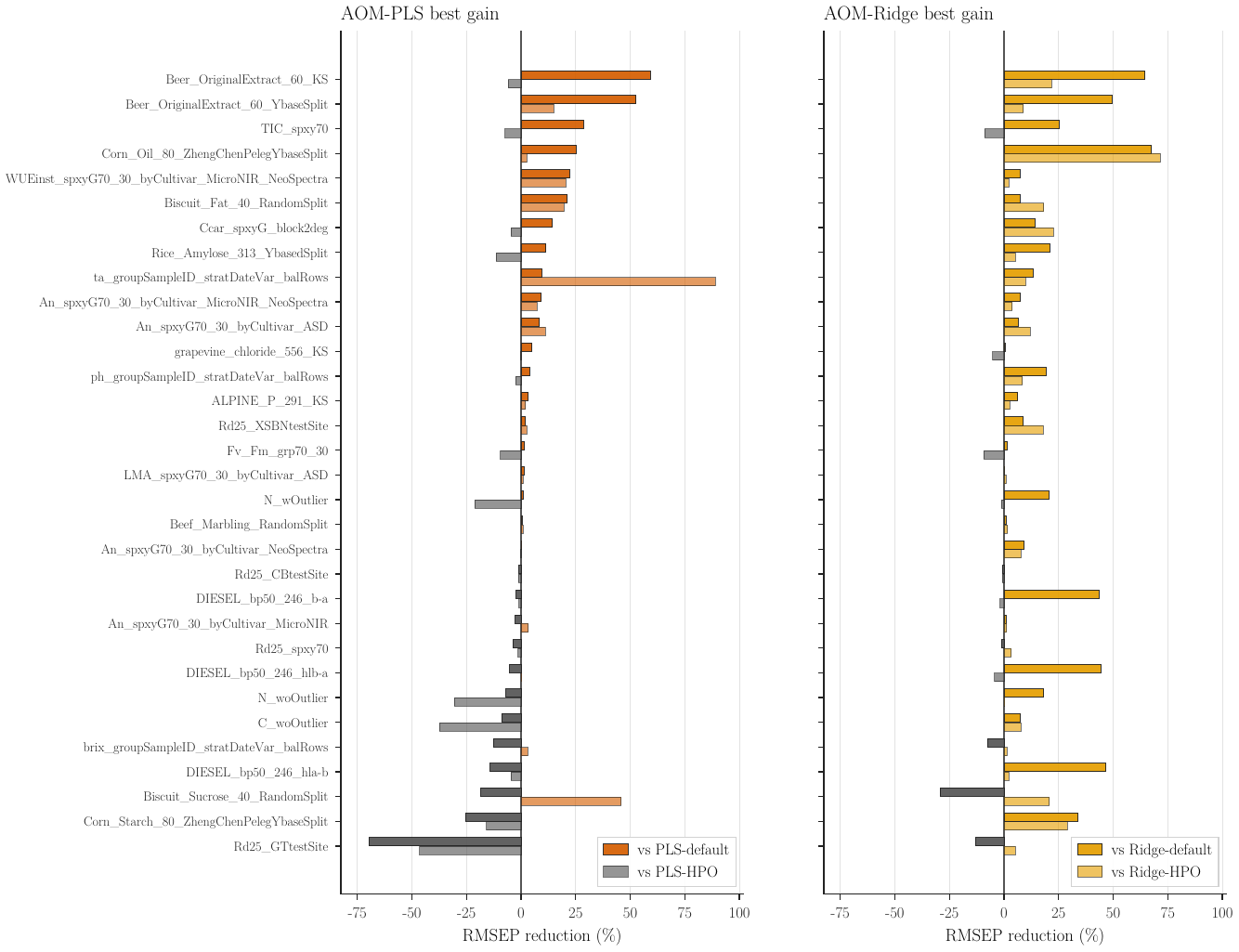}
  \caption{Per-dataset RMSEP reduction for the selected operator-adaptive
  variants against default and HPO references; positive bars favour the
  operator-adaptive model.}
  \label{fig:s_gain}
\end{figure}

The long per-dataset RMSEP table (landscape longtable below) and the per-dataset
ratio heatmap (Figure~\ref{fig:s_heat}) give the full breakdown behind the
aggregate medians, and the gain-per-dataset figure (Figure~\ref{fig:s_gain})
visualizes where the selected operator-adaptive variants help and where they do
not.

\begin{table}[t]
  \centering
  \caption{Per-dataset absolute figures of merit on the strict paired regression
  intersection ($N=32$). RMSEP is in the original response units of each dataset
  (not comparable across rows); $R^2$ and $\mathrm{RPD}=1/\sqrt{1-R^2}$ are for
  AOM-Ridge (simple). A dash marks non-positive $R^2$ --- the failure cases
  discussed in the main-text failure-mode analysis (e.g.\ the An/ASD, Ccar and
  Rd25-site rows). Values are generated by
  \texttt{paper/repro/absolute\_fom.py} from the on-disk run CSVs and reconcile
  with the paired ratios reported in the main text.}
  \label{tab:absolute_fom}
  \scriptsize
  \setlength{\tabcolsep}{3pt}
  \begin{tabularx}{\linewidth}{Xrrrrrrr}
\toprule
Dataset & $n_{\mathrm{test}}$ & \multicolumn{4}{c}{RMSEP (original units)} & AOM-\allowbreak{}Ridge & AOM-\allowbreak{}Ridge \\
\cmidrule(lr){3-6}
 & & AOM-\allowbreak{}PLS & PLS-\allowbreak{}HPO & AOM-\allowbreak{}Ridge & Ridge-\allowbreak{}HPO & $R^2$ & RPD \\
\midrule
ALPINE\_\allowbreak{}P\_\allowbreak{}291\_\allowbreak{}KS & 44 & 0.0605 & 0.0622 & 0.0603 & 0.0583 & 0.643 & 1.67 \\
An\_\allowbreak{}spxyG70\_\allowbreak{}30\_\allowbreak{}byCultivar\_\allowbreak{}ASD & 34 & 3.973 & 4.175 & 3.928 & 4.096 & -0.203 & - \\
An\_\allowbreak{}spxyG70\_\allowbreak{}30\_\allowbreak{}byCultivar\_\allowbreak{}MicroNIR & 35 & 3.802 & 4.06 & 3.663 & 3.555 & 0.187 & 1.11 \\
An\_\allowbreak{}spxyG70\_\allowbreak{}30\_\allowbreak{}byCultivar\_\allowbreak{}MicroNIR\_\allowbreak{}NeoSpectra & 35 & 3.807 & 4.206 & 4.021 & 3.984 & 0.021 & 1.01 \\
An\_\allowbreak{}spxyG70\_\allowbreak{}30\_\allowbreak{}byCultivar\_\allowbreak{}NeoSpectra & 37 & 5.108 & 5.04 & 4.687 & 4.78 & -0.214 & - \\
Beef\_\allowbreak{}Marbling\_\allowbreak{}RandomSplit & 278 & 73.51 & 74.09 & 72.13 & 73.09 & 0.472 & 1.38 \\
Beer\_\allowbreak{}OriginalExtract\_\allowbreak{}60\_\allowbreak{}KS & 20 & 0.2746 & 0.2209 & 0.2422 & 0.2593 & 0.977 & 6.61 \\
Beer\_\allowbreak{}OriginalExtract\_\allowbreak{}60\_\allowbreak{}YbaseSplit & 20 & 0.2918 & 0.2987 & 0.3032 & 0.2977 & 0.985 & 8.05 \\
Biscuit\_\allowbreak{}Fat\_\allowbreak{}40\_\allowbreak{}RandomSplit & 32 & 0.4735 & 0.5428 & 0.2004 & 0.5759 & 0.990 & 9.87 \\
Biscuit\_\allowbreak{}Sucrose\_\allowbreak{}40\_\allowbreak{}RandomSplit & 32 & 1.18 & 2.385 & 1.053 & 1.521 & 0.927 & 3.70 \\
C\_\allowbreak{}woOutlier & 1209 & 2.578 & 1.773 & 1.761 & 2.031 & 0.604 & 1.59 \\
Ccar\_\allowbreak{}spxyG\_\allowbreak{}block2deg & 3229 & 68.32 & 55.95 & 60.05 & 66.68 & -559.326 & - \\
Corn\_\allowbreak{}Oil\_\allowbreak{}80\_\allowbreak{}ZhengChenPelegYbaseSplit & 16 & 0.0294 & 0.0259 & 0.0262 & 0.0676 & 0.977 & 6.62 \\
Corn\_\allowbreak{}Starch\_\allowbreak{}80\_\allowbreak{}ZhengChenPelegYbaseSplit & 16 & 0.1392 & 0.1552 & 0.1091 & 0.1673 & 0.981 & 7.31 \\
DIESEL\_\allowbreak{}bp50\_\allowbreak{}246\_\allowbreak{}b-\allowbreak{}a & 113 & 3.178 & 3.189 & 3.425 & 2.96 & 0.947 & 4.35 \\
DIESEL\_\allowbreak{}bp50\_\allowbreak{}246\_\allowbreak{}hla-\allowbreak{}b & 113 & 3.035 & 3.055 & 2.492 & 2.739 & 0.979 & 6.86 \\
DIESEL\_\allowbreak{}bp50\_\allowbreak{}246\_\allowbreak{}hlb-\allowbreak{}a & 113 & 3.05 & 3.089 & 2.842 & 2.681 & 0.964 & 5.24 \\
Fv\_\allowbreak{}Fm\_\allowbreak{}grp70\_\allowbreak{}30 & 167 & 0.0312 & 0.0283 & 0.031 & 0.028 & 0.306 & 1.20 \\
LMA\_\allowbreak{}spxyG70\_\allowbreak{}30\_\allowbreak{}byCultivar\_\allowbreak{}ASD & 472 & 0.3032 & 0.3073 & 0.3028 & 0.3051 & 0.915 & 3.43 \\
N\_\allowbreak{}wOutlier & 1207 & 0.3581 & 0.2884 & 0.3202 & 0.2696 & 0.952 & 4.59 \\
N\_\allowbreak{}woOutlier & 1207 & 0.3223 & 0.24 & 0.2712 & 0.2251 & 0.966 & 5.41 \\
Rd25\_\allowbreak{}CBtestSite & 146 & 0.2288 & 0.2282 & 0.2255 & 0.231 & 0.364 & 1.25 \\
Rd25\_\allowbreak{}GTtestSite & 173 & 0.1978 & 0.2241 & 0.1953 & 0.2303 & 0.322 & 1.21 \\
Rd25\_\allowbreak{}XSBNtestSite & 151 & 0.2809 & 0.2852 & 0.2595 & 0.2909 & 0.195 & 1.11 \\
Rd25\_\allowbreak{}spxy70 & 141 & 0.1789 & 0.1834 & 0.1741 & 0.1815 & 0.551 & 1.49 \\
Rice\_\allowbreak{}Amylose\_\allowbreak{}313\_\allowbreak{}YbasedSplit & 110 & 2.338 & 2.048 & 1.936 & 2.162 & 0.863 & 2.70 \\
TIC\_\allowbreak{}spxy70 & 19 & 4.065 & 3.343 & 3.862 & 3.124 & 0.677 & 1.76 \\
WUEinst\_\allowbreak{}spxyG70\_\allowbreak{}30\_\allowbreak{}byCultivar\_\allowbreak{}MicroNIR\_\allowbreak{}NeoSpectra & 35 & 1.518 & 1.754 & 1.465 & 1.48 & 0.250 & 1.15 \\
brix\_\allowbreak{}groupSampleID\_\allowbreak{}stratDateVar\_\allowbreak{}balRows & 699 & 4.331 & 4.447 & 3.621 & 3.816 & 0.561 & 1.51 \\
grapevine\_\allowbreak{}chloride\_\allowbreak{}556\_\allowbreak{}KS & 167 & 977.5 & 957.5 & 964.1 & 935 & 0.584 & 1.55 \\
ph\_\allowbreak{}groupSampleID\_\allowbreak{}stratDateVar\_\allowbreak{}balRows & 489 & 0.3434 & 0.3307 & 0.3586 & 0.3251 & 0.395 & 1.29 \\
ta\_\allowbreak{}groupSampleID\_\allowbreak{}stratDateVar\_\allowbreak{}balRows & 489 & 2.016 & 17.77 & 2.021 & 1.977 & 0.319 & 1.21 \\
\bottomrule
\end{tabularx}

\end{table}

\clearpage
\begin{landscape}
\begingroup
\small
\setlength{\tabcolsep}{3.0pt}
\renewcommand{\arraystretch}{1.05}
\begin{longtable}{L{8.0cm}L{9.5cm}rr}
\caption{Per-dataset regression results on the strict intersection $N_{\cap}$ used in the paper.  Relative RMSEP is divided by PLS-standard on the same dataset when the denominator is finite.}\\
\toprule
Dataset & Variant & RMSEP & Rel. RMSEP \\
\midrule
\endfirsthead
\toprule
Dataset & Variant & RMSEP & Rel. RMSEP \\
\midrule
\endhead
ALPINE\_\allowbreak{}P\_\allowbreak{}291\_\allowbreak{}KS & PLS-\allowbreak{}standard-\allowbreak{}numpy & 0.05969 & 1.000 \\
ALPINE\_\allowbreak{}P\_\allowbreak{}291\_\allowbreak{}KS & AOM-\allowbreak{}compact-\allowbreak{}cv5-\allowbreak{}numpy & 0.06079 & 1.018 \\
ALPINE\_\allowbreak{}P\_\allowbreak{}291\_\allowbreak{}KS & ASLS-\allowbreak{}AOM-\allowbreak{}compact-\allowbreak{}cv5-\allowbreak{}numpy & 0.0606 & 1.015 \\
ALPINE\_\allowbreak{}P\_\allowbreak{}291\_\allowbreak{}KS & AOM-\allowbreak{}default-\allowbreak{}nipals-\allowbreak{}adjoint-\allowbreak{}numpy & 0.06733 & 1.128 \\
ALPINE\_\allowbreak{}P\_\allowbreak{}291\_\allowbreak{}KS & nirs4all-\allowbreak{}AOM-\allowbreak{}PLS-\allowbreak{}default & 0.06733 & 1.128 \\
ALPINE\_\allowbreak{}P\_\allowbreak{}291\_\allowbreak{}KS & FastAOM-\allowbreak{}sparse-\allowbreak{}chains-\allowbreak{}supervised & 0.05844 & 0.979 \\
ALPINE\_\allowbreak{}P\_\allowbreak{}291\_\allowbreak{}KS & FastAOM-\allowbreak{}sparse-\allowbreak{}chains-\allowbreak{}compact & 0.05843 & 0.979 \\
ALPINE\_\allowbreak{}P\_\allowbreak{}291\_\allowbreak{}KS & FastAOM-\allowbreak{}single-\allowbreak{}chain-\allowbreak{}compact & 0.05846 & 0.979 \\
\bottomrule
\end{longtable}
\endgroup
\end{landscape}
\clearpage

\section{Classification details}
\label{sec:s_classification}
Table~\ref{tab:s_clsfull} gives the full classification comparison (balanced
accuracy), and Table~\ref{tab:s_clscalib} adds the probability-calibration view:
median balanced accuracy, log-loss and expected calibration error (ECE) per
variant. Operator-adaptive PLS-DA improves balanced accuracy (0.625 vs 0.452) and
log-loss (1.16 vs 1.39) over PLS-DA but is less well-calibrated (ECE 0.32 vs
0.11) --- more accurate and confident, yet over-confident. The operator-adaptive
Ridge classifiers improve balanced accuracy with calibration on par with PLS-DA
(ECE 0.09--0.11, log-loss 0.66--0.81). PLS-DA is promoted in the main text purely
for the methodological point that covariance-space selection transfers to
classification; the Ridge classifiers are the better-calibrated option.

\begin{table}[t]
  \centering
  \caption{Full classification results (balanced accuracy).}
  \label{tab:s_clsfull}
  \small
  \begin{tabularx}{\linewidth}{Xrrrr}
\toprule
Comparison & $N$ & Median $\Delta$ balanced acc. & 95\% CI & Wins; $p_{\mathrm{Holm}}$ \\
\midrule
AOM-PLS-DA-global-simpls-covariance vs PLS-DA & 13 & 0.159 & 0.129--0.422 & 12/13; 0.007 \\
POP-PLS-DA-simpls-covariance vs PLS-DA & 13 & 0.052 & 0.035--0.275 & 11/13; 0.009 \\
AOM-PLS-DA-global-nipals-adjoint vs PLS-DA & 13 & 0.030 & 0.000--0.111 & 8/13; 0.105 \\
POP-PLS-DA-nipals-adjoint vs PLS-DA & 13 & -0.037 & -0.098--0.048 & 5/13; 0.682 \\
AOMRidgeCls-global-compact vs PLS-DA & 14 & 0.175 & 0.145--0.264 & 13/14; 8.5e-04 \\
AOMRidgeCls-branch\_global-compact vs PLS-DA & 14 & 0.169 & 0.137--0.272 & 13/14; 9.2e-04 \\
AOMRidgeCls-superblock-compact vs PLS-DA & 14 & 0.165 & 0.124--0.243 & 14/14; 5.5e-04 \\
AOMRidgeCls-active-compact vs PLS-DA & 14 & 0.163 & 0.118--0.228 & 14/14; 5.5e-04 \\
\bottomrule
\end{tabularx}

\end{table}

\begin{table}[t]
  \centering
  \caption{Classification probability calibration: median balanced accuracy,
  log-loss and expected calibration error (ECE) across datasets and seeds.}
  \label{tab:s_clscalib}
  \small
  \begin{tabularx}{\linewidth}{Xrrr}
\toprule
Variant & Median bal.\ acc. & Median log-loss & Median ECE \\
\midrule
PLS-\allowbreak{}DA-\allowbreak{}standard & 0.452 & 1.388 & 0.114 \\
AOM-\allowbreak{}PLS-\allowbreak{}DA-\allowbreak{}global-\allowbreak{}simpls-\allowbreak{}covariance & 0.625 & 1.160 & 0.320 \\
AOM-\allowbreak{}PLS-\allowbreak{}DA-\allowbreak{}global-\allowbreak{}nipals-\allowbreak{}adjoint & 0.517 & 1.345 & 0.107 \\
\midrule
AOMRidgeCls-\allowbreak{}global-\allowbreak{}compact & 0.613 & 0.793 & 0.112 \\
AOMRidgeCls-\allowbreak{}superblock-\allowbreak{}compact & 0.617 & 0.680 & 0.091 \\
AOMRidgeCls-\allowbreak{}branch\_global-\allowbreak{}compact & 0.609 & 0.805 & 0.110 \\
AOMRidgeCls-\allowbreak{}active-\allowbreak{}compact & 0.605 & 0.663 & 0.091 \\
\bottomrule
\end{tabularx}

\end{table}

\section{Seed and split sensitivity}
\label{sec:s_seed}
Tables~\ref{tab:s_seed}--\ref{tab:s_seedaudit} report seed stability and the
deterministic-split audit for the Ridge selector variants. Table~\ref{tab:s_determinism}
gives the per-variant across-seed RMSEP variability over seeds 0/1/2. The
operator-adaptive \emph{Ridge} calibrations are essentially deterministic
(per-dataset RMSEP standard deviation $0$ to machine precision, $0$ of $25$
datasets varying), so seed jitter is not a meaningful robustness axis for them;
the operator-adaptive \emph{PLS} models do vary on a fraction of datasets
($40$--$48$ of $53$ for the cross-validated selectors) because the inner
cross-validation fold assignment uses the seed. In all cases the held-out test
\emph{partition} is fixed across seeds (the $n_{\text{test}}$ count is invariant),
so this measures selection-CV stability, not robustness to redrawing the test set
--- which is the held-out-site transfer reported in the main text.

\begin{table}[t]
  \centering
  \caption{Seed-stability summary.}
  \label{tab:s_seed}
  \small
  \begin{tabularx}{\linewidth}{p{0.16\linewidth}Xrrrr}
\toprule
Family & Variant & Full seeds & Mean $\rho$ & Seed CV & Winner changes \\
\midrule
AOM-\allowbreak{}PLS & PLS-\allowbreak{}standard & 32 & 1.000 & 0.000 & 14 \\
AOM-\allowbreak{}PLS & AOM-\allowbreak{}compact-\allowbreak{}cv5 & 32 & 0.994 & 0.017 & 14 \\
AOM-\allowbreak{}PLS & ASLS-\allowbreak{}AOM-\allowbreak{}compact-\allowbreak{}cv5 & 32 & 0.994 & 0.021 & 14 \\
AOM-\allowbreak{}PLS & AOM-\allowbreak{}default & 32 & 1.000 & 0.000 & 14 \\
AOM-\allowbreak{}Ridge top5 & Ridge-\allowbreak{}raw & 23 & 1.000 & 0.000 & 0 \\
AOM-\allowbreak{}Ridge top5 & AOMRidge global compact & 23 & 1.000 & 0.000 & 0 \\
AOM-\allowbreak{}Ridge top5 & AOMRidge local knn50 & 23 & 1.000 & 0.000 & 0 \\
AOM-\allowbreak{}Ridge top5 & AOMRidge local blended & 23 & 1.000 & 0.000 & 0 \\
\bottomrule
\end{tabularx}

\end{table}

\begin{table}[t]
  \centering
  \caption{Ridge deterministic-split (seed) audit.}
  \label{tab:s_seedaudit}
  \small
  \begin{tabularx}{\linewidth}{Xrrrr}
\toprule
Variant & Audit datasets & Seeds 0/1/2 complete & Max RMSEP span across seeds & Datasets with non-zero span \\
\midrule
AOMRidge-\allowbreak{}AutoSelect & 18 & 18 & 0 & 0 \\
AOMRidge-\allowbreak{}Blender & 18 & 18 & 0 & 0 \\
\bottomrule
\end{tabularx}

\end{table}

\begin{table}[t]
  \centering
  \caption{Across-seed RMSEP variability (seeds 0/1/2): median and maximum
  per-dataset standard deviation, and the number of datasets with any variation.
  AOM-Ridge variants are deterministic; AOM-PLS cross-validated selectors vary
  with the inner-CV seed.}
  \label{tab:s_determinism}
  \begingroup
  \tiny
  \setlength{\tabcolsep}{2pt}
  \renewcommand{\arraystretch}{0.72}
  \begin{tabularx}{\linewidth}{lXrrrr}
\toprule
Family & Variant & \#Datasets\,/\,\#Seeds & Median seed std & Max seed std & \#Datasets varying \\
\midrule
AOM-\allowbreak{}PLS & AOM-\allowbreak{}compact-\allowbreak{}cv3-\allowbreak{}numpy & 53\,/\,3 & 0.0156 & 1692.4391 & 48/53 \\
 & AOM-\allowbreak{}compact-\allowbreak{}cv5-\allowbreak{}numpy & 53\,/\,3 & 0.0076 & 310.6608 & 45/53 \\
 & POP-\allowbreak{}simpls-\allowbreak{}covariance-\allowbreak{}numpy & 53\,/\,3 & 0.0072 & 244.1130 & 34/53 \\
 & ASLS-\allowbreak{}AOM-\allowbreak{}compact-\allowbreak{}cv5-\allowbreak{}numpy & 53\,/\,3 & 0.0057 & 169.1556 & 40/53 \\
 & POP-\allowbreak{}nipals-\allowbreak{}adjoint-\allowbreak{}numpy & 53\,/\,3 & 0.0038 & 80.2668 & 32/53 \\
 & AOM-\allowbreak{}default-\allowbreak{}nipals-\allowbreak{}adjoint-\allowbreak{}numpy & 53\,/\,3 & 0 & $<$1e$-$9 & 0/53 \\
 & AOM-\allowbreak{}compact-\allowbreak{}simpls-\allowbreak{}covariance-\allowbreak{}numpy & 53\,/\,3 & 0 & $<$1e$-$9 & 0/53 \\
 & PLS-\allowbreak{}standard-\allowbreak{}numpy & 53\,/\,3 & 0 & $<$1e$-$9 & 0/53 \\
 & nirs4all-\allowbreak{}AOM-\allowbreak{}PLS-\allowbreak{}default & 53\,/\,3 & 0 & $<$1e$-$9 & 0/53 \\
\midrule
AOM-\allowbreak{}Ridge & AOMRidge-\allowbreak{}Local-\allowbreak{}compact-\allowbreak{}cv-\allowbreak{}blended & 25\,/\,3 & 0 & $<$1e$-$9 & 0/25 \\
 & AOMRidge-\allowbreak{}Local-\allowbreak{}compact-\allowbreak{}knn50 & 25\,/\,3 & 0 & $<$1e$-$9 & 0/25 \\
 & AOMRidge-\allowbreak{}global-\allowbreak{}compact-\allowbreak{}none-\allowbreak{}split\_aware & 25\,/\,3 & 0 & $<$1e$-$9 & 0/25 \\
 & Ridge-\allowbreak{}raw & 25\,/\,3 & 0 & $<$1e$-$9 & 0/25 \\
\midrule
Tuned PLS & PLS-\allowbreak{}HPO (25 trials) & 36\,/\,3 & 0.0640 & 32.6476 & 36/36 \\
\midrule
Tuned Ridge & Ridge-\allowbreak{}HPO (60 trials) & 35\,/\,3 & 0.0478 & 18.5849 & 35/35 \\
\midrule
AOM-\allowbreak{}Ridge-\allowbreak{}Cls (bal. acc.) & AOMRidgeCls-\allowbreak{}branch\_global-\allowbreak{}compact & 14\,/\,3 & 0.0108 & 0.0597 & 11/14 \\
 & AOMRidgeCls-\allowbreak{}global-\allowbreak{}compact & 14\,/\,3 & 0.0103 & 0.0614 & 11/14 \\
 & AOMRidgeCls-\allowbreak{}active-\allowbreak{}compact & 14\,/\,3 & 0.0050 & 0.0282 & 11/14 \\
 & AOMRidgeCls-\allowbreak{}superblock-\allowbreak{}compact & 14\,/\,3 & 0.0035 & 0.0417 & 11/14 \\
\midrule
AOM-\allowbreak{}PLS-\allowbreak{}DA (bal. acc.) & POP-\allowbreak{}PLS-\allowbreak{}DA-\allowbreak{}nipals-\allowbreak{}adjoint & 13\,/\,3 & 0.0372 & 0.0898 & 13/13 \\
 & POP-\allowbreak{}PLS-\allowbreak{}DA-\allowbreak{}simpls-\allowbreak{}covariance & 13\,/\,3 & 0.0278 & 0.1619 & 13/13 \\
 & AOM-\allowbreak{}PLS-\allowbreak{}DA-\allowbreak{}global-\allowbreak{}nipals-\allowbreak{}adjoint & 13\,/\,3 & 0 & $<$1e$-$9 & 0/13 \\
 & PLS-\allowbreak{}DA-\allowbreak{}standard & 16\,/\,3 & 0 & $<$1e$-$9 & 0/16 \\
 & AOM-\allowbreak{}PLS-\allowbreak{}DA-\allowbreak{}global-\allowbreak{}simpls-\allowbreak{}covariance & 13\,/\,3 & 0 & $<$1e$-$9 & 0/13 \\
\bottomrule
\end{tabularx}

  \endgroup
\end{table}

\section{Two-sided test sensitivity}
\label{sec:s_twosided}
\paragraph{Reporting convention (p-value family).} All $p$-values reported in
the main text and here are computed over the \emph{full} family of paired
comparisons (one-sided Wilcoxon signed-rank favouring the row method, Holm-corrected
within that displayed family). A smaller \emph{pre-registered} family of seven
comparisons yields more permissive values for the same data --- for example
AOM-Ridge Blender vs Ridge-HPO is $p_{\mathrm{Holm}}=0.00308$ under the
pre-registered family but $0.033$ under the full family, and AOM-Ridge
AutoSelect vs Ridge-HPO is $0.044$ versus $0.741$. We report the
\textbf{conservative full-family values throughout}; the pre-registered values
are not used. The AOM-Ridge-Local ($k$NN-50) one-sided $p=1.000$ reflects that it
is a \emph{worse} method (median ratio $1.212$); under the full-family two-sided
test it is significantly worse ($p=0.002$), consistent with its role as an
intentional negative example.

Table~\ref{tab:s_paired} repeats the full-family paired comparisons and adds an
explicit two-sided sensitivity column: alongside the one-sided $p_{\mathrm{Holm}}$
(Wilcoxon favouring the row method, Holm-corrected over the displayed family) we
tabulate the two-sided $p_{\mathrm{Holm}}$ obtained by re-running the same Holm
correction over two-sided Wilcoxon $p$-values for the identical 23-comparison
family. The two-sided values are, as expected, roughly a factor of two larger for
the directional AOM-Ridge advantages---AOM-Ridge Blender vs Ridge-default stays
significant at $p_{\mathrm{Holm}}=5.2\times10^{-4}$ and vs Ridge-HPO moves to
$0.063$, while AOM-Ridge global-compact vs Ridge-default remains significant at
$0.013$---so the headline conclusions are unchanged under the two-sided
convention. The AOM-Ridge-Local ($k$NN-50) row, whose one-sided $p_{\mathrm{Holm}}=1.000$
merely reflects that the test was run in the wrong direction for an intentionally
worse method, becomes $0.002$ two-sided within this full family, i.e.\
significantly worse.

\begin{table}[t]
  \centering
  \caption{Paired statistics (sensitivity).}
  \label{tab:s_paired}
  \begingroup
  \tiny
  \setlength{\tabcolsep}{2pt}
  \renewcommand{\arraystretch}{0.72}
  \begin{tabularx}{\linewidth}{Xrrrrrr}
\toprule
Comparison & $N$ & Median ratio & 95\% CI & Wins & One-sided $p_{\mathrm{Holm}}$ & Two-sided $p_{\mathrm{Holm}}$ \\
\midrule
ASLS-\allowbreak{}AOM-\allowbreak{}compact-\allowbreak{}cv5 vs PLS-\allowbreak{}standard & 32 & 0.973 & 0.934--0.995 & 22/32 & 0.184 & 0.350 \\
AOM-\allowbreak{}compact-\allowbreak{}cv5 vs PLS-\allowbreak{}standard & 32 & 0.983 & 0.963--1.005 & 20/32 & 0.518 & 0.937 \\
AOM-\allowbreak{}default-\allowbreak{}nipals-\allowbreak{}adjoint vs PLS-\allowbreak{}standard & 32 & 1.003 & 0.966--1.049 & 16/32 & 1.000 & 1.000 \\
ASLS-\allowbreak{}AOM-\allowbreak{}compact-\allowbreak{}cv5 vs PLS-\allowbreak{}default & 32 & 0.985 & 0.916--1.018 & 20/32 & 1.000 & 1.000 \\
AOM-\allowbreak{}compact-\allowbreak{}cv5 vs PLS-\allowbreak{}default & 32 & 0.991 & 0.970--1.000 & 22/32 & 0.896 & 1.000 \\
AOM-\allowbreak{}default-\allowbreak{}nipals-\allowbreak{}adjoint vs PLS-\allowbreak{}default & 32 & 1.005 & 0.975--1.057 & 15/32 & 1.000 & 1.000 \\
ASLS-\allowbreak{}AOM-\allowbreak{}compact-\allowbreak{}cv5 vs PLS-\allowbreak{}HPO & 32 & 1.002 & 0.977--1.035 & 15/32 & 1.000 & 1.000 \\
AOM-\allowbreak{}compact-\allowbreak{}cv5 vs PLS-\allowbreak{}HPO & 32 & 0.990 & 0.975--1.021 & 19/32 & 1.000 & 1.000 \\
PLS-\allowbreak{}HPO vs PLS-\allowbreak{}default & 32 & 0.992 & 0.939--1.010 & 19/32 & 1.000 & 1.000 \\
Ridge-\allowbreak{}HPO vs Ridge-\allowbreak{}default & 32 & 0.962 & 0.913--1.004 & 19/32 & 1.000 & 1.000 \\
AOMRidge-\allowbreak{}global-\allowbreak{}compact-\allowbreak{}none vs Ridge-\allowbreak{}default & 32 & 0.974 & 0.892--0.993 & 25/32 & 0.007 & 0.013 \\
AOMRidge-\allowbreak{}Blender vs Ridge-\allowbreak{}default & 32 & 0.918 & 0.808--0.937 & 27/32 & 2.6e-04 & 5.2e-04 \\
AOMRidge-\allowbreak{}Blender vs Ridge-\allowbreak{}HPO & 32 & 0.966 & 0.918--0.985 & 25/32 & 0.033 & 0.063 \\
AOMRidge-\allowbreak{}AutoSelect vs Ridge-\allowbreak{}HPO & 32 & 0.963 & 0.929--0.996 & 22/32 & 0.741 & 1.000 \\
AOMRidge-\allowbreak{}global-\allowbreak{}compact-\allowbreak{}none vs Ridge-\allowbreak{}HPO & 32 & 0.984 & 0.934--1.020 & 19/32 & 1.000 & 1.000 \\
AOMRidge-\allowbreak{}Local-\allowbreak{}knn50 vs Ridge-\allowbreak{}HPO & 23 & 1.212 & 1.051--1.550 & 4/23 & 1.000 & 0.002 \\
FastAOM-\allowbreak{}sparse-\allowbreak{}chains-\allowbreak{}supervised vs PLS-\allowbreak{}standard & 32 & 0.953 & 0.903--0.980 & 23/32 & 0.741 & 1.000 \\
FastAOM-\allowbreak{}sparse-\allowbreak{}chains-\allowbreak{}compact vs PLS-\allowbreak{}standard & 32 & 0.953 & 0.909--0.979 & 22/32 & 0.518 & 0.937 \\
FastAOM-\allowbreak{}single-\allowbreak{}chain-\allowbreak{}compact vs PLS-\allowbreak{}standard & 32 & 0.999 & 0.977--1.036 & 16/32 & 1.000 & 1.000 \\
FastAOM-\allowbreak{}sparse-\allowbreak{}chains-\allowbreak{}supervised vs PLS-\allowbreak{}HPO & 32 & 0.993 & 0.931--1.070 & 16/32 & 1.000 & 1.000 \\
FastAOM-\allowbreak{}sparse-\allowbreak{}chains-\allowbreak{}compact vs PLS-\allowbreak{}HPO & 32 & 0.995 & 0.919--1.066 & 16/32 & 1.000 & 1.000 \\
FastAOM-\allowbreak{}single-\allowbreak{}chain-\allowbreak{}compact vs PLS-\allowbreak{}HPO & 32 & 1.066 & 1.021--1.131 & 7/32 & 1.000 & 0.937 \\
FastAOM-\allowbreak{}sparse-\allowbreak{}chains-\allowbreak{}supervised vs ASLS-\allowbreak{}AOM-\allowbreak{}compact-\allowbreak{}cv5 & 32 & 1.021 & 0.975--1.067 & 15/32 & 1.000 & 1.000 \\
\bottomrule
\end{tabularx}

  \endgroup
\end{table}

\section{Source-family clustered sensitivity}
\label{sec:s_sourcefam}
The paired tests treat each dataset as independent, but the 32 datasets of the
strict intersection come from only 15 source families (e.g.\ GRAPEVINE\_LeafTraits
contributes six, DarkResp four, and BERRY, DIESEL and COLZA three each), so
dataset-level inference risks pseudo-replication. We therefore collapse the
per-dataset RMSEP ratios to a single median per family and re-run a one-sided
sign test on the 15 family-level values, with a 95\% confidence interval from a
cluster bootstrap that resamples \emph{families} rather than datasets
(Table~\ref{tab:s_sourcefam}). The principal conclusions strengthen rather than
weaken: AOM-Ridge (global-compact) and the AOM-Ridge Blender beat default Ridge in
$14/15$ and $13/15$ families (family-median ratios $0.974$ and $0.868$; sign
$p=4.9\times10^{-4}$ and $3.7\times10^{-3}$), and AOM-PLS (compact-cv5) beats
default PLS in $13/15$ families ($p=3.7\times10^{-3}$), because collapsing
correlated within-family rows removes their dilution. The comparisons that were
already marginal after Holm correction at the dataset level---ASLS-AOM vs
PLS-default and the AOM-Ridge Blender vs the \emph{tuned} Ridge-HPO---fall to
$p=0.059$ at the family level (the latter's cluster CI, $0.916$--$1.019$, now
spans one), so they are reported as suggestive once pseudo-replication is taken
into account.

\begin{table}[t]
  \centering
  \caption{Source-family clustered sensitivity. Per-dataset RMSEP ratios are
  collapsed to per-family medians (32 datasets $\to$ 15 families); the family-level
  column gives the median ratio with a 95\% cluster-bootstrap CI (resampling
  families) and the one-sided family sign test.}
  \label{tab:s_sourcefam}
  \small
  \begin{tabularx}{\linewidth}{Xrrrrl}
\toprule
 & \multicolumn{2}{c}{Cohort} & \multicolumn{2}{c}{Median RMSEP ratio} & Family-level \\
\cmidrule(lr){2-3}\cmidrule(lr){4-5}
Comparison & Datasets & Families & Dataset-level & Family-level (95\% CI) & wins / sign $p$ \\
\midrule
AOMRidge-\allowbreak{}global-\allowbreak{}compact-\allowbreak{}none vs Ridge-\allowbreak{}default & 32 & 15 & 0.974 & 0.974 (0.762--0.995) & 14/15, $p=4.9\times10^{-4}$ \\
AOMRidge-\allowbreak{}Blender vs Ridge-\allowbreak{}default & 32 & 15 & 0.918 & 0.868 (0.788--0.987) & 13/15, $p=0.004$ \\
AOM-\allowbreak{}compact-\allowbreak{}cv5 vs PLS-\allowbreak{}default & 32 & 15 & 0.991 & 0.974 (0.961--0.999) & 13/15, $p=0.004$ \\
ASLS-\allowbreak{}AOM-\allowbreak{}compact-\allowbreak{}cv5 vs PLS-\allowbreak{}default & 32 & 15 & 0.985 & 0.968 (0.950--0.996) & 11/15, $p=0.059$ \\
AOMRidge-\allowbreak{}Blender vs Ridge-\allowbreak{}HPO & 32 & 15 & 0.966 & 0.970 (0.916--1.019) & 11/15, $p=0.059$ \\
\bottomrule
\end{tabularx}

\end{table}

\section{Failure modes}
\label{sec:s_failure}
Table~\ref{tab:s_failure} lists the datasets where the operator-adaptive models
trail the preprocessing-HPO winner, dominated by fitted scatter or baseline
corrections (SNV, MSC, ASLS), together with the non-finite rows excluded by
construction.

\begin{table}[t]
  \centering
  \caption{Failure modes and non-finite rows.}
  \label{tab:s_failure}
  \small
  \begin{tabularx}{\linewidth}{Xrrrr}
\toprule
Dataset & Variants logged & Finite RMSEP rows & Min RMSEP & Max RMSEP \\
\midrule
FinalScore\_\allowbreak{}grp70\_\allowbreak{}30\_\allowbreak{}scoreQ & 9 & 0 & n/a & n/a \\
Tleaf\_\allowbreak{}grp70\_\allowbreak{}30 & 9 & 0 & n/a & n/a \\
Brix\_\allowbreak{}spxy70 & 9 & 9 & 0.9499 & 2.0182 \\
C\_\allowbreak{}woOutlier & 9 & 9 & 1.6018 & 2.7590 \\
LMA\_\allowbreak{}spxyG70\_\allowbreak{}30\_\allowbreak{}byCultivar\_\allowbreak{}ASD & 9 & 9 & 0.3032 & 0.7767 \\
N\_\allowbreak{}wOutlier & 9 & 9 & 0.3429 & 1.3923 \\
N\_\allowbreak{}woOutlier & 9 & 9 & 0.3134 & 1.3955 \\
Rice\_\allowbreak{}Amylose\_\allowbreak{}313\_\allowbreak{}YbasedSplit & 9 & 9 & 1.8873 & 5.3238 \\
brix\_\allowbreak{}groupSampleID\_\allowbreak{}stratDateVar\_\allowbreak{}balRows & 9 & 9 & 4.1629 & 5.3942 \\
\bottomrule
\end{tabularx}

\end{table}

\section*{Declaration of Generative AI and AI-assisted technologies in the writing process}
During the preparation of this work, the authors used Anthropic Claude and
OpenAI Codex to assist with code review, implementation, repository
organization, benchmark aggregation scripts, LaTeX editing and drafting or
revision of explanatory text.  After using these tools, the authors reviewed,
edited and verified the code, numerical results, references and manuscript
content as needed, and take full responsibility for the content of the
publication.

\bibliographystyle{unsrtnat}
\bibliography{references}